%% file: 00_main.tex
\documentclass{article}
\usepackage{jfrExamplee}
\usepackage{graphicx}
\usepackage{apalike}
\usepackage{setspace}
\usepackage{appendix}
\usepackage{todonotes}
\usepackage{hyperref}
\usepackage{algorithmic}
\usepackage[linesnumbered,nofillcomment,commentsnumbered]{algorithm2e}

\title{Skyeye Team at MBZIRC 2020: A team of aerial and ground robots for GPS-denied autonomous fire extinguishing in an urban building scenario}

\author{
Simón Martínez-Rozas$^1$$^*$ \\
\texttt{simon.martinez@uantof.cl} \\ 
\And
Rafael Rey$^1$$^*$ \\
\texttt{rreyarc@upo.es} \\
\And 
David Alejo$^1$$^*$ \\
\texttt{daletei@upo.es} \\
\And
Domingo Acedo$^1$$^*$ \\
\texttt{dacegom@upo.es} \\
\AND
José Antonio Cobano$^1$ \\
\texttt{jacobsua@upo.es} \\
\And
Alejandro Rodríguez-Ramos$^2$ \\		
\texttt{alejandro.rramos@upm.es} \\ 
\And
Pascual Campoy$^2$ \\
\texttt{pascual.campoy@upm.es} \\
\And
 Luis Merino$^1$ \\ 
 \texttt{lmercab@upo.es} \\
 \And
 Fernando Caballero$^1$ \\
\texttt{fcaballero@us.es} \\
 \AND
$^1$Service Robotics Laboratory \\
Universidad Pablo de Olavide    \\
Crta. Utrera, km. 1, 41013, Sevilla, Spain \\
$^*$\emph{Equal contribution}
\AND
$^2$ Computer Vision and Aerial Robotics Group \\
Centre for Automation and Robotics (C.A.R.)\\ 
Universidad Politécnica de Madrid (UPM-CSIC)\\
Calle Jose Gutierrez Abascal 2, 28006 Madrid, Spain
}

\newcommand{\revisionSo}[1]{\textcolor{black}{#1}}
\newcommand{\revisionSr}[1]{\textcolor{black}{#1}}
\newcommand{\revisionFer}[1]{\textcolor{blue}{#1}} 
\newcommand{\revisionJac}[1]{\textcolor{black}{#1}}
\newcommand{\revisionDat}[1]{\textcolor{black}{#1}}
\newcommand{\revisionLuis}[1]{\textcolor{black}{#1}}

\begin{document}

\maketitle

\begin{abstract}
The paper \revisionSo{presents a} 
framework for fire extinguishing in an urban scenario by a team of aerial and ground robots. The system was developed \revisionLuis{to address Challenge 3} 
of the 2020 Mohamed Bin Zayed International Robotics Challenge (MBZIRC). The challenge required to autonomously detect, locate and extinguish fires \revisionSo{on} 
different floors of a building, as well as in its surroundings. The multi-robot system developed consists of a heterogeneous robot team of up to three Unmanned Aerial Vehicles (UAV) and one Unmanned Ground Vehicle (UGV). \revisionDat{We describe the} main hardware and software components for UAV and UGV platforms \revisionDat{and also present} the main algorithmic components of the system: a 3D LIDAR-based mapping and localization module able to work in GPS-denied scenarios;  a global planner and a fast local re-planning system for robot navigation; infrared-based perception and robot actuation control for fire extinguishing; and a mission executive and coordination module based on Behavior Trees. The paper finally describes the results obtained during \revisionSo{the} competition, where the system worked fully autonomously and scored in all the trials performed. \revisionLuis{The presented system ended in 7th position out of 20 teams in the Challenge 3 competition and in 5th position (out of 17 teams) in the Challenge 3 entry to the Grand Finale (Grand Challenge) of MBZIRC 2020 competition}. 
\end{abstract}

\section{Introduction}
\label{sec:Introduction}
\input{01_Introduction.tex}


\section{Hardware}
\label{sec:Hardware}
\input{02_Hardware}

\section{3D Mapping and Localization}
\label{sec:mapping}
\input{03_MapAndLocation.tex}

\section{Autonomous robot navigation}
\label{sec:navigation}
\input{04_Navigation.tex}

\section{Fire perception and extinguishing}
\label{sec:fire}
\input{05_FireExtinguishing.tex}

\section{Mission Executive and Multi-Robot Coordination}
\label{sec:executive}

\input{06_ExecutiveLayer}

\section{Experiments and Results}
\label{sec:experiments}
\input{07_ExperimentAndResults.tex}


\section{Conclusions and Lessons Learnt}
\label{sec:conclusion}
\input {10_ConclusionsNew2.tex}

\subsubsection*{Acknowledgments}
This work has been supported by the Spanish Ministry of Science, Innovation and Universities (COMCISE RTI2018-100847-B-C22, MCIU/AEI/FEDER, UE) and the Khalifa University under contract no. 2020-MBZIRC-10. The work of S. Martinez-Rozas is partially supported by the Antofagasta University.

UPO would like to acknowledge Paulo Alvito and the IDMind\footnote{https://www.idmind.pt} company for the support with the ground robot hardware, Marcel Richter and the AeroGuillena airfield for the support during the tests, and Danilo Tardioli, from the University of Zaragoza, Spain, for his multi-master ROS software for multi-robot teams.

\bibliographystyle{apalike}
\bibliography{dataset}

%



\end{document}

%% file: 01_Introduction.tex
The application of robotic technologies in disaster management and rescue operations is increasing in the last years due to the advances in the design of platforms and the mapping, localization, perception, planning, and coordination capabilities \cite{delmerico2019}. The goal is to decrease the risks that humans face in such activities and, at the same time, increase the effectiveness of the operations. 

Robot competitions are spreading in the world to foster advancements \revisionSo{in robotics research } 
for these applications and to allow proper benchmarking between different approaches. Some examples related to search and rescue and disaster management scenarios are: the Defense Advanced Research Projects Agency (DARPA) Robotics Challenge \cite{pratt2013}, RoboCup Rescue competition \cite{sheh2016}, euRathlon and ERL Emergency \cite{winfield2017eurathlon}, DARPA Fast Lightweight Autonomy program \cite{mohta2018}, Trinity College Robot Competition Schedule Changes \cite{trinity2020} and the Mohamed Bin Zayed International Robotics Challenge (MBZIRC) \cite{mbzirc2020}.

Among these 
missions and challenges, \revisionSo{firefighting } 
is seen as a relevant application of robot systems given the dangerous conditions in which firefighters have to operate. Firefighters are at \revisionSo{a} constant risk of being burned, becoming trapped, or inhaling smoke. The risks of losing human lives could be greatly reduced by using autonomous robots capable of searching, detecting and extinguishing fires \revisionSo{\cite{RongjieChen2019}}.
In general, current automatic firefighting systems are fixed. For instance, automatic fire sprinklers and alarms are used in heavily populated and hazardous areas for rapidly extinguishing any fire. Mobile robotic firefighting systems such as ground and aerial robots, on the other hand, could travel into unsafe and/or inaccessible areas to collect information through \revisionSo{onboard} 
sensors (visual and infrared cameras, LIDAR, etc) \revisionSo{\cite{PECHO2019461}} and even to perform extinguishing operations \revisionSo{\cite{INNOCENTE201980}}. This would offer critical new capabilities on real-time situation awareness and remote extinguishing to \revisionSo{firefighters}
. Achieving this vision still requires, though, advancing the state of the art and robustness of current robots both in software and hardware, as well as in different directions \revisionSo{ \cite{delmerico2019}, as performance in both indoors and outdoors spaces, clear fire image detection \revisionSo{ \cite{drones3010017}} and capability of autonomous navigation\cite{Imdoukh2017}.} 

\revisionLuis{In this regard, } the MBZIRC 2020 competition included a Challenge 3 denoted: ``Team of Robots to Fight Fire in High Rise Building''. In it, a team of robots needed to collaborate to detect, localize and extinguish fires in a simulated high-rise building. This involved putting out fires located inside and on the walls of a building at different heights and floors, and outside the building, by using water and fire blankets. The mission is challenging  and requires to broaden the state-of-the-art in several areas: deployment and coordination of multiple heterogeneous robots (UAVs and UGV) in the same urban workspace; autonomous indoor/outdoor navigation in cluttered environments; autonomous search and location of fires; hardware adaptations and control for fire extinguishing.

This paper describes in detail a multi-robot fully autonomous system conceived for fire extinguishing in such urban high rise building firefighting scenario. The system is composed \revisionSo{of three }
Unmanned Aerial Vehicles (UAV) and one Unmanned Ground Vehicle (UGV). The system ended in 7th position out of 20 teams in the Challenge 3 \revisionLuis{competition} of MBZIRC 2020, and \revisionLuis{in 5th position (out of 17 teams) in the Challenge 3 entry to the Grand Finale (Grand Challenge), in which this Challenge 3 and the other two Challenges of the competition were run simultaneously}\footnote{\url{https://www.mbzirc.com/winning-teams/2020}}.


The main contributions of the paper are related directly \revisionLuis{to the technical advances required to address MBZIRC Challenge 3 by a fleet of heterogeneous robot in fully autonomous mode}.
First of all, a multi-modal robust localization module, able to operate in GPS-denied scenarios. While the competition allowed the use of GPS and even DGPS (with a penalization in this latter case), the ability to operate without GPS was crucial to navigate autonomously and score in all trials, given the intermittent reception due to satellite occlusion and other factors. Secondly, a flexible mission composition and execution system based on Behaviour Trees (BT) \cite{TRO17Colledanchise} 
in charge of combining all modules to define the behaviors of each robot and that of the whole team to accomplish missions. The versatility of BTs was fundamental to quickly adapt to the conditions of the competition. Finally, the paper also contributes to the robotics community with the description of the specialized navigation, control and perception modules, and hardware employed in the competition.

The paper is \revisionSo{organized} 
into eight sections. Section \ref{sec:Hardware} describes the hardware designs for the UAV and UGV platforms. Section \ref{sec:mapping} presents the mapping and localization system implemented. The autonomous navigation system of each platform is described in Section \ref{sec:navigation}. Section \ref{sec:fire} shows the approach proposed to detect and extinguish the fires and Section \ref{sec:executive} presents the framework used to carry out the mission. Experiments are presented in Section \ref{sec:experiments}. Finally, lessons learned and conclusions are described in Section \ref{sec:conclusion}.

\subsection{State of the art}

The idea of using robots for fire fighting has been present in the community for many years. Early designs can be traced back to \cite{bardshaw91}, which describes the functional and mechanical design of a ground robot for fire detection and first intervention in indoor environments. In \cite{amano01}, the authors show the design of a climbing robot for helping fire activities in buildings. All in all, robots are considered \revisionSo{a} 
promising tool to fight fires because of their ability to reach inaccessible zones, assist in dangerous environments, \revisionLuis{and} to carry sensors to monitor and control the fires. 

Actually, robotic technology is being often deployed as part of firefighting operations in the last years. Current examples can be found in the Notre Dame fire in 2019 with the firefighting ground robot called Collosus\footnote{\url{https://www.shark-robotics.com/colossus}}. It is a remote-controlled firefighting automaton designed and built by Shark Robotics. TC800-FF is another remotely operated ground robot with some autonomous navigation capabilities designed to assist fire-fighter during operations\footnote{\url{https://www.robotpompier.com/en/}}. Two of the most popular firefighting ground robots are the Thermite RS1-T3 and RS2-T2\footnote{\url{http://www.roboticfirefighters.com/}}. The aim of all these robots is to replace fighters and to keep them safe and free from the heavy work that distracts them and takes time away from solving problems quickly and effectively. Most of these systems are mainly teleoperated. While teleoperation can be enough for certain scenarios, the inclusion of autonomous capabilities \revisionSo{expands} 
further the utility of such systems, \revisionLuis{and can be very relevant when fleets of several robots are involved, like in the MBZIRC challenge}. 

Furthermore, in the case of high-rise buildings and in other scenarios, the use of aerial robots can be beneficial. In the last 15 years, Unmanned Aerial Vehicles (UAVs) have been more and more used in wildfire fighting (e.g. \cite{ambrosia2003demonstrating,casbeer2006cooperative,merino2006cooperative,skeele2016aerial,bailon2018planning}). These systems are mainly used as ``eyes in the skies", to provide situational awareness to first response teams, given their capability of carrying sensors and position in vantage points. Some of these works \cite{casbeer2006cooperative,merino2006cooperative,bailon2018planning} have shown the advantages of using teams of cooperating UAVs for fire fighting. In particular, we have also shown in our former works \cite{merino2006cooperative,merino2012unmanned} the added value of cooperating heterogeneous UAVs with different capabilities and that can carry complementary sensors in order to localize precisely fire spots and discard false alarms. Currently, drones are used as part of wildfire operations, as for instance during the Australian bushfires where drones were used to search and rescue koalas affected by the fires \revisionLuis{\cite{koala2020}}.

Those works do not consider the inclusion of fire suppression mechanisms, which for UAVs is also a challenge. A novel hose type robot, which can fly directly into the fire source via a water-jet, has been proposed in \cite{ando2018}. In \cite{yamaha2016} the concept of attaching a drone to the tip of a fire hose and remotely extinguishing a fire is presented. These studies have not been tested in a real fire extinguishing task. Problems such as the payload and influence of the tension of the hose itself on the flight must be addressed to bring the UAV’s hose closer to the fire source. 

When considering the use of autonomous robots to fight urban fires in buildings \revisionLuis{as in MBZIRC}, additional challenges are present, like the \revisionLuis{navigation in cluttered and GPS-denied environments or with limited GPS coverage}, as is typically the case in urban canyons and/or close to and inside buildings, and the operation in low-visibility conditions \cite{schneider2017using}. The last years have seen important advances in the navigation of UAVs in GPS-denied scenarios (for instance \cite{shen2014multi,mohta2018,Perezgrau18_3,usenko2020tum}). \revisionLuis{However, there are still issues that need to be addressed to navigate and precisely position the UAVs to attack fires in urban scenarios like the one posed by Challenge 3 of MBZIRC.}

The cooperation between UAVs and UGVs presents additional challenges. In \cite{michael2014collaborative} the synergies between a UGV and UAVs are exploited to map buildings affected by earthquakes. \revisionSo{But there are not many works in the literature that consider fully-autonomous ground and aerial robots for fire fighting in urban environments}. In \cite{maza2011experimental}, a system of multiple UAVs cooperating with a static sensor network for a fire scenario in a building is presented. UAVs are used to deploy new sensors and provide situation awareness, which is then used to attack the fire by firefighters. The work shows again how cooperation can be very valuable. \revisionLuis {However,} the system assumes GPS coverage, no ground robots are considered, and no extinguishing is performed by the robots themselves. 


\revisionLuis{In sum, addressing a fire extinguishing scenario in a high-rise building by a team of robots, as the one posed by MBZIRC, still requires advances in localization, navigation, perception, and mission coordination and execution of the system as a whole. The paper presents in the next sections the proposed advances in those lines, discussing further literature pertinent for each case}.

\subsection{Problem statement}
\label{sec:problem_statement}

\begin{figure}[!tb]
	\centering
    \includegraphics[width =0.9\textwidth]{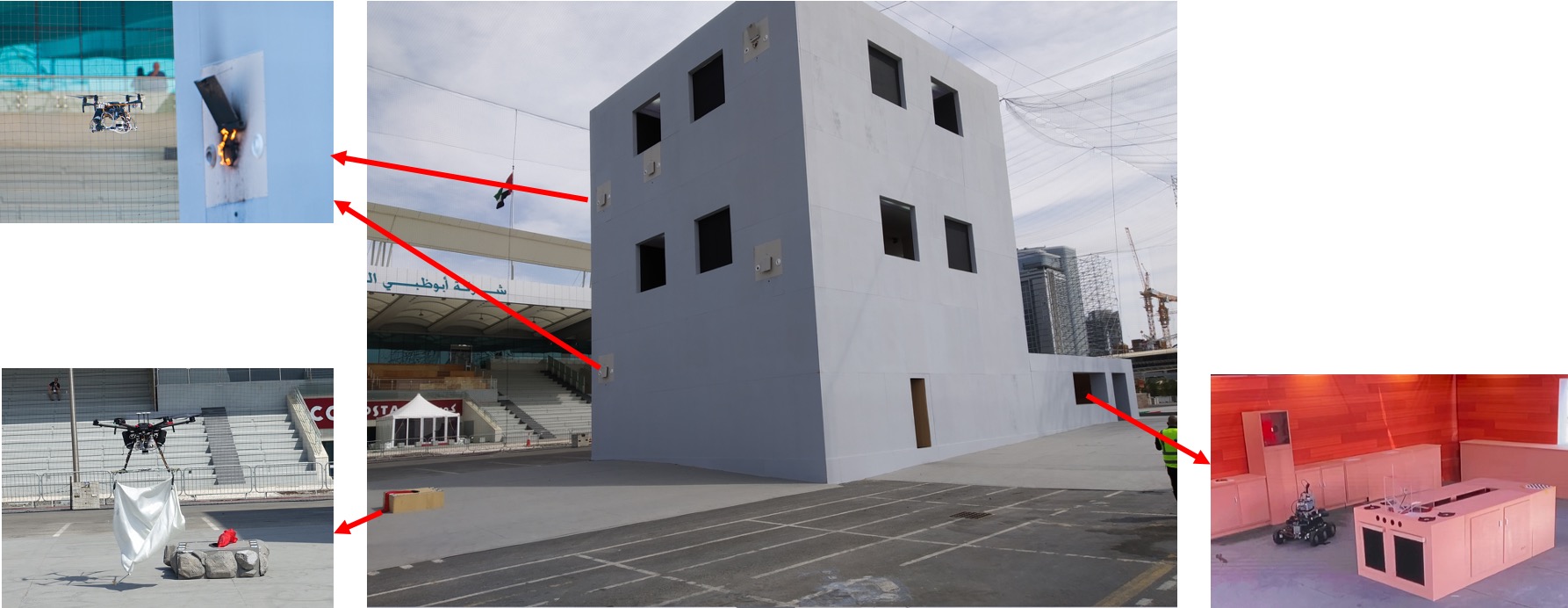}
	\caption{Arena of Challenge 3. There are simulated fires on the facades of the building (top-left), and inside the building (down, right), located on the ground, first and second floor (to be suppressed with water). There are also fires located on the ground outside (down, left) that need to be suppressed with a blanket.}
	\label{fig:arena}
\end{figure}

This section summarizes the main aspects of the rules, conditions, and scoring of the MBZIRC Challenge 3. 
The challenge requires a team of UAVs (maximum three) and UGV (maximum one) collaborating in order to autonomously extinguish a series of simulated fires in an urban high rise building firefighting scenario\footnote{\url{https://www.mbzirc.com/challenge/2020}}. Figure \ref{fig:arena} shows the arena for Challenge 3, with a size of approximately that of half a football pitch. A 15-meter high structure simulates the building. 

The robots have to extinguish a set of 6 fires located in the 3 floors of the building using water. For each floor, there is one fire inside the building and one fire on the facade (see Fig. \ref{fig:arena}). Furthermore, there are 2 fires located outside the building at ground level that must be extinguished using fire blankets carried by the robots. The locations of the fires are changed randomly between trials.

Each fire is scored between 0 and 1. For the fires in the building the maximum score is obtained if the amount of water properly thrown into the fire is 1 liter. Simulated fire spots on the ground score when they are covered with a blanket (with a maximum when they are totally covered). The final score is obtained by a weighted sum for all 8 fires. Different fires receive different weights according to their difficulty\footnote{\url{https://www.mbzirc.com/scoring-scheme}}.

The mission can be attempted in either fully autonomous or manual mode. The mission is considered in manual mode from the moment there is an intervention by a human 
(the only interventions allowed are to replenish water and change batteries and blankets). Any score in manual mode is considered below a score in autonomous mode, or to disambiguate in case of \revisionSr{a} draw for scores in autonomous mode. GPS is allowed. RTK/DGPS can be used although a penalty of 25\% will be applied if it is used. 
The duration of Challenge 3 is 15 minutes, 
with a 5-minute preparation slot. 

%% file: 02_Hardware.tex


The robot team to address the challenge consists of one 
UGV and and up to 3 UAVs. They are adequately equipped to perform navigation and to detect and extinguish the fire for this firefighting scenario.

\subsection{Unmanned ground vehicle}

The ground platform used is called SIAR \cite{siar2020}, a 6-wheeled robot with independent traction system and a pan\&tilt mechanism on top. The robot is able to navigate at a maximum speed of 0.7 m/s and it has battery autonomy for more than 4 hours of operation (motion and sensing/computation). The robot integrates an i7 computer, local sensing and a 3D LIDAR. The main sensors and mechanisms  used in the ground platform are (see Fig. \ref{fig:siar_sensors}):

\begin{figure}[!t]
	\centering
    \includegraphics[width = 1.00\textwidth]{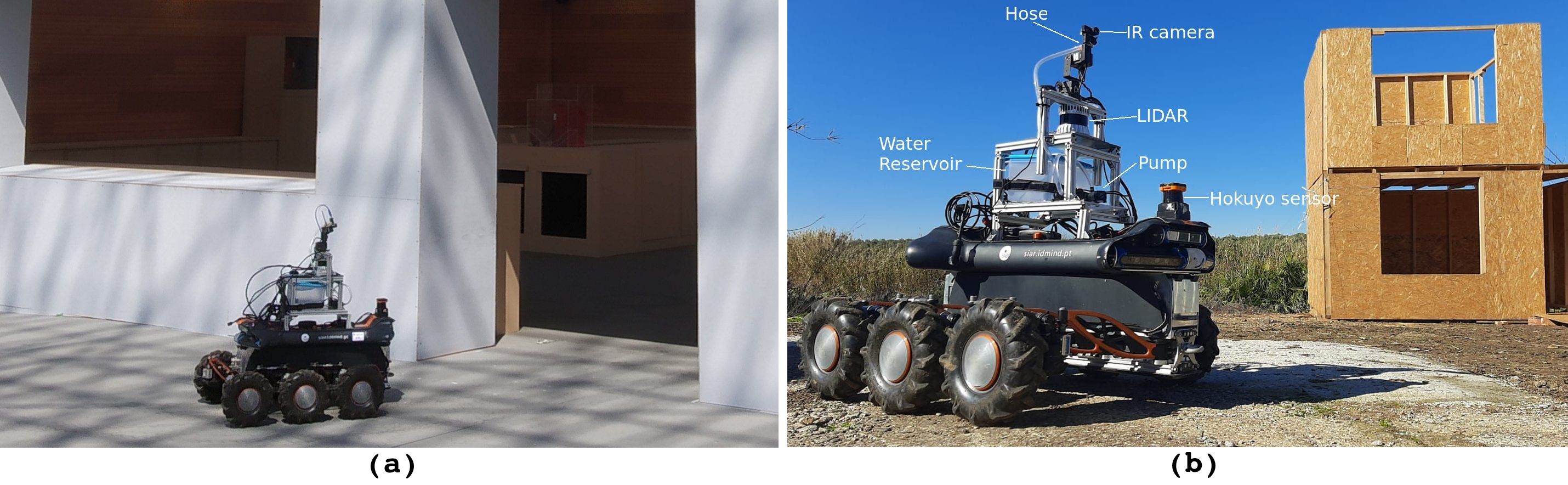}
	\caption{The ground platform used for Challenge 3.(a) Showing performance in the scenario of MBZIRC (b) Deploy of onboard sensors and Water Extinguishing System}
	\label{fig:siar_sensors}
\end{figure}

\begin{itemize}

    \item GPS sensor UBlox 8 and ArduIMUv3 Inertial Measurement Unit (IMU).
    \item LIDARs: A 3D OS-1-16 LIDAR manufactured by Ouster has been installed on top of the robot. The LIDAR is able to provide 320k 3D points per second, ranging from 0.2 to 120 meters. The information gathered by this sensor is used for localization and mapping. Furthermore, a Hokuyo UTM-30LX 2D sensor is installed. The information of this sensor is used for local obstacle detection/avoidance.
    \item Cameras: A SeekThermal IR camera to detect the sources of heat and an Orbbec's ASTRA camera that provides the operator with images to operate the platform if needed. 
    \item Pan\&Tilt unit: It moves the thermal camera for searching and centering the fire, and at the same time, it holds and guides the hose of the extinguishing system to aim the fire.

    \item Extinguishing System: It consists in a series connection of two water pumps (5V, 4.8 W, 300cm Hmax, 300L/H, no valve), one MOSFET module 5V, hoses, and a 3-liter water tank (within the limits of the competition). To improve the intake of water, the tank is placed above the water pumps. Alike, the output hose is placed above the pump to prevent the water from going out, as the pumps do not have a valve. 
     With this, we have a system that ejects water 2 meters in a straight line from the hose outlet
\end{itemize}

\subsection{Unmanned aerial vehicles}

Regarding to the aerial platforms, the models used for the competition are adaptations of the DJI Matrice 210 V2 and the DJI Matrice 600 Pro (see Fig. \ref{fig:m600}). We selected these platforms because of the robustness of the hardware and software, support that the provider offers, and also their facility to communicate the PC on-board with the Robot Operating System (ROS) \cite{Quigley09}. The drone's structure was adapted in order to perform the tasks and to integrate new hardware. For that, we take into account the maximum weight that can carry each drone and also the dimension constraints of the competition. Each UAV robot integrates an i7 NUC computer and local sensing. The main sensors and mechanisms added to the UAVs are (see Fig. \ref{fig:m600}):

\begin{figure}[!t]
	\centering
    \includegraphics[width =  0.84\textwidth]{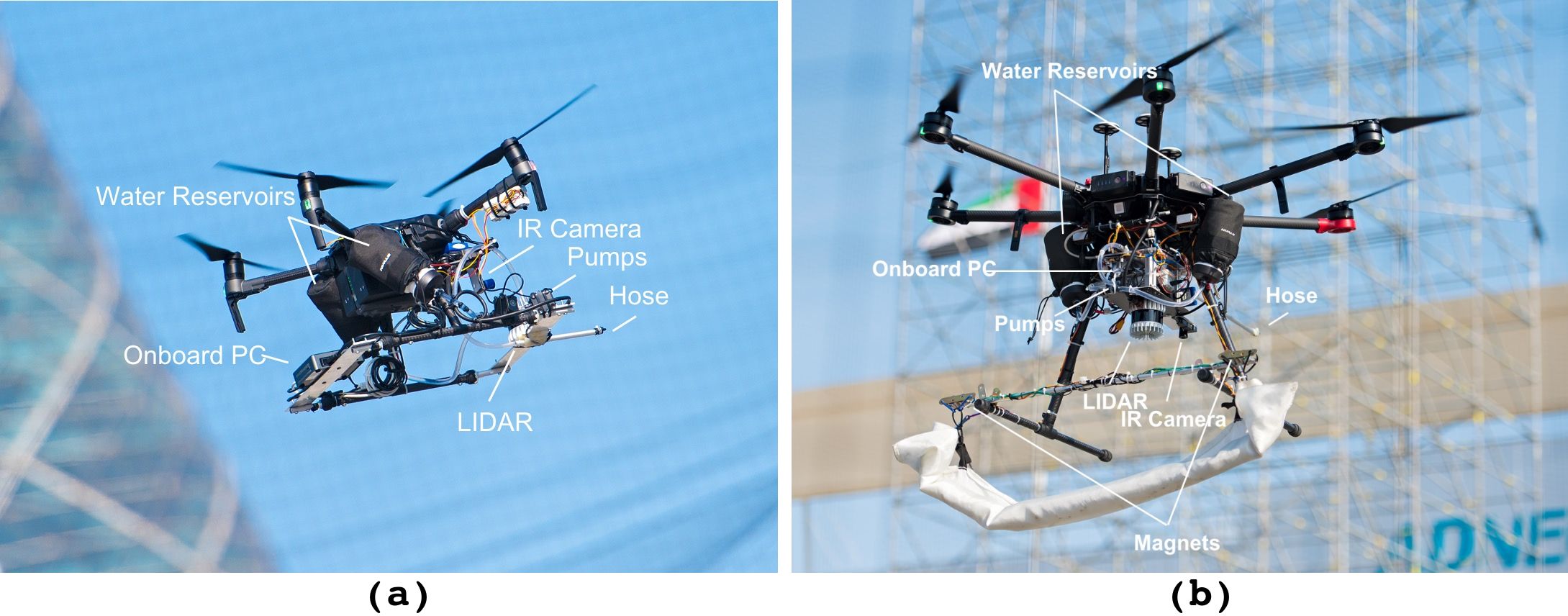}
	\caption{(a) Adapted DJI Matrice 210v2 for facade fire extinguishing, with LIDAR and PC placed to avoid blocking the down-looking sensors. (b) Adapted DJI Matrice 600 for blanket deployment and facade fire extinguishing.}
	\label{fig:m600}
\end{figure}

\begin{itemize}
    \item DJI Matrice onboard sensors. This includes single GPS, Inertial Measurement Unit (IMU) and barometric altimeter. DJI Matrice 600 intergrates 3 GPS receivers and 3 IMUs for redundancy and fault detection purposes, but they behave as single devices. 
    \item 3D LIDAR. OS-1-16 LIDAR by Ouster. The purpose of this sensor is threefold: environment mapping, localization, and odometry. It is the basis for the generation of the 3D map of the environment, and to compute an accurate robot odometry for the UAVs. It is also used as main sensor for robot localization in all robots, and to detect obstacles by the navigation module. 
    \item Infrared camera: SeekThermal whose images are used for fire detection. The M210 carries one camera looking forward, while the M600 carries an additional one in a nadir configuration for the ground fires.
    \item Water Extinguishing System: we use the same pumps as the UGV platform, 
    but with less capacity due to payload constraints. We use two 0.5-liter water tank, which are positioned on each leg of the drone to evenly distribute their weight
    . The outlet of the tanks are joined in a coupling that carries the water to the pumps. These are located in a structure below the drone and the tanks. An aluminum tube is used to guide the outlet hose from the pumps to a safe area in which the air turbulence generated by the drone itself does not disperse the water when it is ejected. A flow restrictor is placed at the outlet of the hose to increase the pressure and range. 
    \item Blanket Extinguishing system (M600 only): we use six electromagnets to carry and drop the blanket. The electromagnets are located on a bar that joins the back of the two legs of the drone. The blanket is transported rolled up to prevent the air turbulence generated by the drone from producing a sail effect. The electromagnets are distributed into two groups of three electromagnets each. In each group, there is one electromagnet for rolling the blanket and two for dropping it. When desired, the rolling electromagnets are deactivated and the blanket is unrolled, and when the blanket is to be thrown, the drop electromagnets are deactivated. 
\end{itemize}

%% file: 03_MapAndLocation.tex
\revisionLuis{The application considered requires that the robots are localized} seamlessly and accurately both in GPS and GPS-denied environments (including buildings interiors, urban canyons and, in general, areas with high dilution of precision), with a smooth transition among them. \revisionLuis{Our solution considers a map-based localization system able to fuse data from different sensors.}


The mapping and localization system proposed consists of two main components
: a LIDAR-based odometry and mapping and an online map-based global localization module. These systems create the global map during the mapping phase, and estimate the global position of the robot into the map during the mission respectively. 
While there are differences in the implementations for the UAVs and UGV, the modules are based on the same principles.

\subsection{3D Mapping and multi-sensor odometry}
\label{sec:3d_mapping}

\revisionFer{Different approaches can be used for 3D mapping and odometry in aerial vehicles. Stereo cameras \cite{Kitt2010,Geiger2011,Schmid2013} and RGBD \cite{Endres2012,Kerl2013} are common choices indoors or for outdoor navigation close to obstacles. However RGBD and time-of-flight cameras cannot work under strong sunlight, and stereo cameras would need a large baseline in order to accurately estimate distances at 25 meters approximately (as the case of Challenge 3). Bundle adjustment approaches \cite{CUCCI20171,pagliari2015reviewer1} can be used offline in combination to visual cameras or short-baseline stereo to build accurate maps outdoors, but they cannot provide reliable odometry during the robot operation. On the other hand, 3D-LIDAR approaches work outdoors, range up to 120 meters and can perceive the 360º of the robot's environment.}

Thus, the base for the proposed mapping and localization systems is a \revisionFer{LIDAR-based} odometry system that allows to estimate the relative motion of the robot. This module considers IMU and 3D LIDAR measurements in order to compute a reliable 6DoF robot odometry in the environment. It provides short-term aligned scans that can be used for local mapping, and a map of the environment. It integrates the IMU estimation together with a LIDAR-based mapping process to provide global consistency to the mapping process.

The LIDAR-based odometry module \revisionLuis{makes use of the} 
LIDAR Odometry and Mapping (LOAM) system \cite{LOAM,zhang2017low-drift}, which decomposes the original SLAM problem into scan registration and mapping. 
The system computes high-precision motion estimates and maps. 
We make use of an open software implementation of the original LOAM system 
\footnote{\url{https://github.com/HKUST-Aerial-Robotics/A-LOAM}}. 



\begin{figure}[!tb]
	\centering
    \includegraphics[width = 0.9\textwidth]{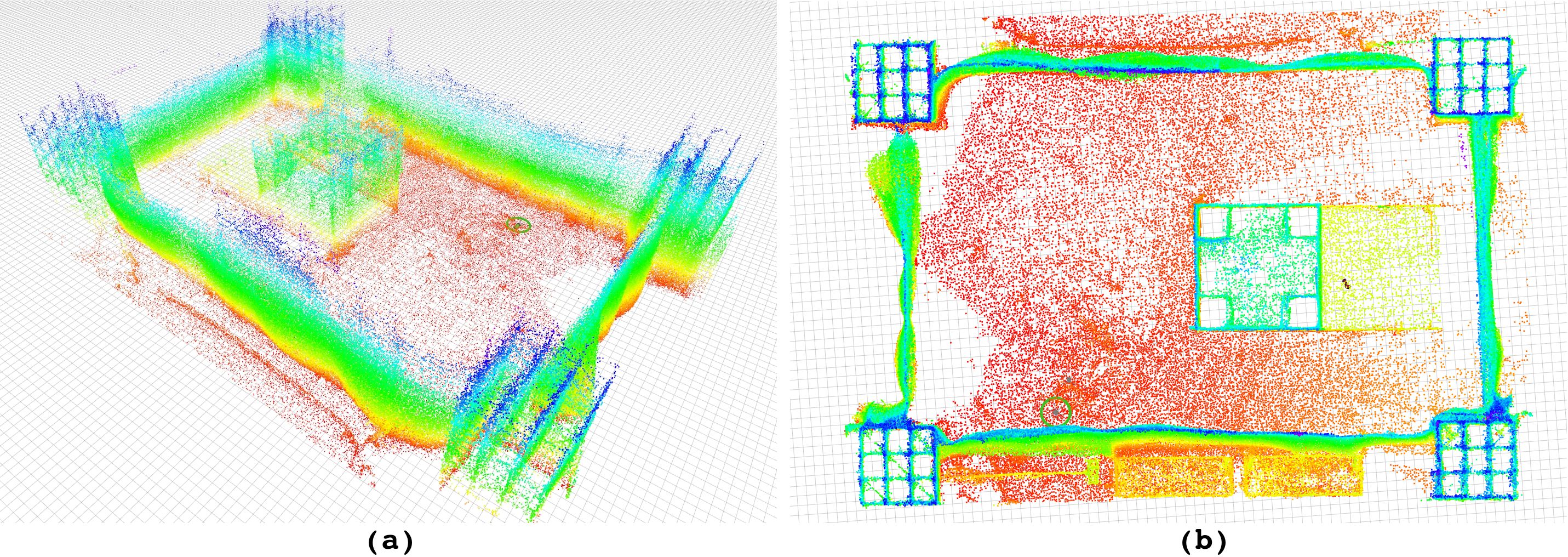}
	\caption{Two views of the map generated for the Challenge 3 of the scenario shown in Fig. \ref{fig:arena} with data provided by the M600 UAV. (a) a map perspective in 3D space and (b) the top view of the map.
	}
	\label{fig:map_example}
\end{figure}

A drone equipped with LIDAR, GPS and on-board sensors (IMU, altimeter, etc) 
was flown manually over the Challenge 3 arena 
, mapping the environment at different altitudes, from 2 meters to the building top. The resulting 3D map is shown in Fig. \ref{fig:map_example}. \revisionFer{It can be seen how the system is able to accurately map the interior of the building and also the four scaffolding structures in the corners. Originally, the map also included the surrounding buildings, but they were removed to reduce the size of the map because these areas were unreachable by the robots.} During the mapping stage, the information from the GPS is used to compute the relative transformation between the map-based global frame and the GPS coordinate system for multi-sensor localization. 

\subsection{Multi-sensor robot localization}
\label{sec:loc} 

\revisionFer{The localization module provides a robust global position of the robot within the map built in Section \ref{sec:3d_mapping}. Map-based approaches for localization are usually better than SLAM approaches in terms of reliability and computational requirements. Thus, decoupling the robot localization from the map building process saves computation resources because the map is built offline once, and the complexity of the algorithm is usually smaller with the corresponding impact in reliability.}

\revisionFer{One of the approaches based on predefined models of the environment is commonly known as teach-replay \cite{Gridseth2019,Royer2007}, which is accomplished in two stages: first the robot is manually piloted along the desired path as in a teaching phase, and an accurate 3D map of the environment is built, along with the robot motion from this learning path; afterwards, this map is used to locate the robot when it repeatedly visits the same path. Although very effective, these approaches are not general enough since they only enables a robot to follow a predetermined trajectory.}

\revisionFer{Monte Carlo Localization (MCL) is another approach that makes use of a known map of the environment, and is commonly used for robot navigation in indoor environments \cite{THRUN200199}. It is a probabilistic localization algorithm that makes use of a particle filter to estimate the pose of the robot within the map, based on sensor measurements. 
An extension of this approach for 6D localization based on 2D laser scanner is presented in \cite{Hornung10}, but it is meant for the 2D motion of humanoid robots in a 3D environment, which makes it not suitable for aerial robots. The use of GPUs can enhance the computing capabilities of the on-board system regarding MCL-based approaches, as proposed by other authors \cite{kanai2015reviewer2,fallon2012reviewer3}.}

\revisionFer{The approach used in this paper is based on Monte Carlo localization. It} extends our previous work presented in \cite{Perezgrau17JARS} with the following aspects:
\begin{itemize}
    \item The particle prediction model has been adapted to absorb non-modelled errors in the odometric system. One of the major issues of MCL approaches are their dependence on the odometric model. A faulty odometry quickly leads to filter divergence due to a wrong dispersion of the particles into the solution space. We included new random terms into the particle prediction model in order to account for these non-modelled errors.
    \item New sensor updates: Compass, altimeter and GPS were integrated into the localization approach in order to build a multi-sensor approach with increased reliability. The  MBZIRC competition requires that the robots are  localized seamlessly and accurately both in GPS and GPS-denied environments, with a smooth transition among them. Integrating multiple sensors improves the reliability of the solution, reducing the dependence on a single sensor and also easy the transition from GPS-denied to full-GPS navigation.
    \item General improvements for faster and robust execution. The localization system is an essential element of the robot navigation, it is involved in almost every robot task, so it must be robust, and efficient because robot on-board computation is very limited. 
\end{itemize}

\noindent The source code of the version used in the experiments is publicly available\footnote{\url{https://github.com/robotics-upo/mcl3d}}.



The localization filter maintains $N$ pose hypotheses (particles) \revisionFer{defined as $\mathbf{p}_{i}^t=[x_i^t, y_i^t, z_i^t, \Psi_i^t]$, where $x_i^t, y_i^t, z_i^t$ and $\Psi_i^t$ refers to the robot position and yaw angle for particle $i$ at time $t$}. Each particle has an associated weight \revisionFer{$w_i^t$} such as \revisionFer{$\Sigma_i^N(w_i^t)=1$}. Note how the robot roll and pitch angles are not included into the robot pose definition. These angles are available and accurate enough in the robot through the use of its onboard IMU. They are fully observable and their values are usually accurate. This greatly reduces the computational complexity of the algorithm, allowing for real-time onboard computation.  

The particles are initialized by setting the initial pose (at the take-off spot) and distributing them in the hypotheses space. The LIDAR-based odometry \revisionFer{described in Section \ref{sec:3d_mapping}} provides increments of the robot pose on the robot frame (after roll and pitch compensation) \revisionFer{$[\Delta x^t, \Delta y^t, \Delta z^t, \Delta \Psi^t]$}, which are used to propagate the distribution of the particles at each time step. Thus, the state of the particles will evolve according to the following expressions for particle \revisionFer{$\mathbf{p}_{i}^t=[x_i^t, y_i^t, z_i^t, \Psi_i^t]$}:

\revisionFer{
\begin{eqnarray}\label{eq:prediction}
    x^{t+1}_{i}&=&x^{t}_{i}+N_x+\Delta x^t \cdot \cos (\Psi^{t}_{i})-\Delta y^t \cdot \sin (\Psi^{t}_{i})\nonumber\\
    y^{t+1}_{i}&=&y^{t}_{i}+N_y+\Delta x^t \cdot \sin (\Psi^{t}_{i})+\Delta y^t \cdot \cos (\Psi^{t}_{i})\nonumber\\
     z^{t+1}_{i}&=&z^{t}_{i}+N_z+\Delta z^t \\
     \Psi^{t+1}_{i}&=&\Psi^{t}_{i}+N_\Psi+\Delta \Psi^t\nonumber
\end{eqnarray}
}

The values of \revisionFer{$\Delta x^t$, $\Delta y^t$, $\Delta z^t$, and $\Delta \Psi^t$} are drawn randomly following a normal distribution centered in their actual values and standard deviations proportional to each increment itself. \revisionFer{The values of $N_x$, $N_y$, $N_z$ and $N_\Psi$ are design parameters, they are also drawn randomly from a zero-centered normal distribution with fixed standard deviation. The main purpose of these parameters are to account for ourliers int the odometry estimation. They force to randomly sample solutions not included into the odometry estimation in view of possible errors. This random noise increases the robustness of the solution by exploring particle states near the odometry prior.} 

The approach evaluates the sensors only when the robot moves above given thresholds in both translation and rotation. When the robot moves above one or both thresholds, it performs an update: using the previous equations to predict robot position according to odometry, updating particles based on sensors and re-sampling the hypotheses space when required. The following sensor readings are used to update the weight \revisionFer{$w_i^t$} associated to each particle's hypothesis \revisionFer{$\mathbf{p}_i^t$}: the point clouds provided by the 3D LIDAR sensor, the 3D position provided by the GPS, the robot roll, pitch and yaw provided by the IMU and the height above the ground provided by the altitude sensor. Each sensor is integrated into the MCL as follows \revisionFer{(for the sake of clarity, the time index $^t$ is removed from the following equations)}:

\begin{itemize}
    \item The point clouds are transformed to each particle pose in order to find correspondences between the cloud and what the map should look like from that particle’s pose. Since this is very expensive computationally, a 3D probability grid is computed offline as in \cite{Hornung10,Perezgrau17Iros}. \revisionFer{Instead of storing binary information about occupancy in the map, this grid stores a value of how likely is that a given position falls within an occupied point of the map. Thus, the probability of each cell $\mathbf{c} = [x_c, y_c, z_c]$ is computed as a Gaussian distribution centered in the closest occupied point in the map  $\mathbf{m}=[x_m,y_m,z_m]$}, and whose variance $\sigma^2$ depends on the sensor noise used in the approach.
    
    \begin{equation}
    grid(\mathbf{c}) = \frac{1}{{ \sqrt {2\pi \sigma^2 } }} e^{{{ - ||\mathbf{c} - \mathbf{m} || ^2 } \mathord{\left/ {\vphantom {{ - ||\mathbf{c} - \mathbf{m} || ^2 } {2\sigma ^2 }}} \right. \kern-\nulldelimiterspace} {2\sigma ^2 }}}
    \end{equation}
    
    Such probability grid only needs to be computed once, it is not required to be updated for a given environment, and relieves from performing numerous distance computations between each cloud point for each particle and its closest occupied point in the map. Besides, each point cloud is first transformed according to the current roll and pitch provided by the on-board IMU. This transformation is done just once per update, reducing the computational requirements as well.
    Then, for every point of the transformed cloud, we access its corresponding value in the 3D probability grid. Such value would be an indicator of how likely is that point to be part of the map. By doing this with every point of the cloud and adding all the probability values, we obtain a figure of how well that particle fits the true location of the aerial robot according to the map.

    Finally, the weight $w_i^{map}$ of each particle $\mathbf{p}_i$ is computed. Assuming that the point cloud is composed of $M$ 3D points $\mathbf{v}_j$, the weight is computed by adding the associated probability grid values:
    \begin{equation}
    \label{eq:wmap}
    w_i^{map}=\frac{1}{M}\sum_{j=1}^{M}grid\mathbf(\mathbf{p}_i(\mathbf{v}_j))
    \end{equation}
    \noindent \revisionFer{where $\mathbf{p}_i(\mathbf{v}_j)$ stands for the transformation of the point $\mathbf{v}_j$ to the particle's state $\mathbf{p}_i$}, and $grid(\mathbf{p}_i(\mathbf{v}_j))$ is the evaluation of the probability grid in such transformed position.
    
    Equation (\ref{eq:wmap}) can be also computed as the product of all $grid(\mathbf{p}_i(\mathbf{v}_j))$. However this was discarded for the following reasons: a) outlier points $\mathbf{v}_j$ might have a significant impact in the weight computation, leading to almost zero in some cases, b) point clouds area easily composed by dozens of thousands of points, given that $grid(\mathbf{p}_i(\mathbf{v}_j))$ ranges from $1$ to $0$, we can fall into numerical errors in the product.
    
    \item The position measurements provided by the GPS are used to check how well each particle 
    matches the sensor reading. First, the GPS position is transformed into the map by means of a known fixed transformation computed during mapping task. Then, a $w^{gps}$ is computed for each particle. Assuming a GPS measurement $\mathbf{p}^{gps}$, the associated weight $w_i^{gps}$ is is computed as follows:
    \begin{equation}
        w_i^{gps}=\frac{1}{{ \sqrt {2\pi \sigma^2 } }} e^{{{ - ||\mathbf{p}_i - \mathbf{p}^{gps} || ^2 } \mathord{\left/ {\vphantom {{ - ||\mathbf{p}_i - \mathbf{p}^{gps} || ^2 } {2\sigma ^2 }}} \right. \kern-\nulldelimiterspace} {2\sigma ^2 }}}
    \end{equation}
    \noindent where the distance computation dismisses the altitude contribution of the GPS, which is subject to significant noise. Also, the standard deviation of the GPS measurement $\sigma$ is computed empirically from the data gathered during the mapping stage and to account for GPS errors in single configuration.
    
    \item The height above the ground is integrated into the MCL through the re-sampling stage. Instead of including the information as a regular update, we decided to include this information into the re-sampling, forcing the hypotheses to be distributed around the altitude provided by the robot altimeter. This way, we can reduce the dispersion of the particles around Z axis, which is easily observable by means of sensors such as laser altimeter, barometer or GPS. 
    
    \item IMU yaw angle is also integrated through the re-sampling stage. The rationale behind this decision is also based on the computational optimization and the nature of the yaw angle. In the experiments area, the yaw angle was checked to be only slightly distorted by the environment, so that we can trust the estimation from the magnetometer. Thus, every time a new re-sampling is performed, the orientation of the particles are drawn from a Gaussian distribution centered in the latest yaw value provided by the IMU. A global to map calibration is used to transform the IMU yaw into the map yaw.
    
\end{itemize}

As mentioned, GPS and point-cloud updates are integrated into the filter. Given the distinct nature of the two technologies involved, we calculate separate weights for each sensing modality. A weighted average is used to obtain the final weight of each particle:
\begin{equation}
w_i=\alpha*w_i^{map}+(1-\alpha)*w_i^{gps}
\label{eq:alpha}
\end{equation}
\noindent where $\alpha$ is chosen depending on the particularities of the indoor environment where the robot is going to operate. If the map used in the MCL does not contain the full environment or its accuracy is not enough to trust the map matching, $\alpha$ should be lower than $0.5$. Whereas if GPS measurements are not accurate, $\alpha$ should be higher.

This approach is used for both robot systems, ground and aerial. However, the dispersion in Z axis is set to a very small value in the predictive model of the ground robot. This relates with the fact that the UGV moves in 2D.


\revisionLuis{\subsection{Experiments}}

\revisionFer{A set of experiments have been conceived in order to validate and benchmark the presented approach. Thus, we first compared the estimation errors of our approach with respect to MCL3D \cite{Perezgrau17JARS}. The software implementation and datasets publicly available here\footnote{https://github.com/fada-catec/amcl3d} have been used to benchmark our approach. The dataset provides the LIDAR data gathered by an aerial robot flying indoors in a 20x20x5 meters volume. The dataset also provides the ground-truth robot position computed with a motion tracking system with millimetric accuracy.} 

\begin{figure}[!t]
    \centering
    \includegraphics[width=0.47\textwidth]{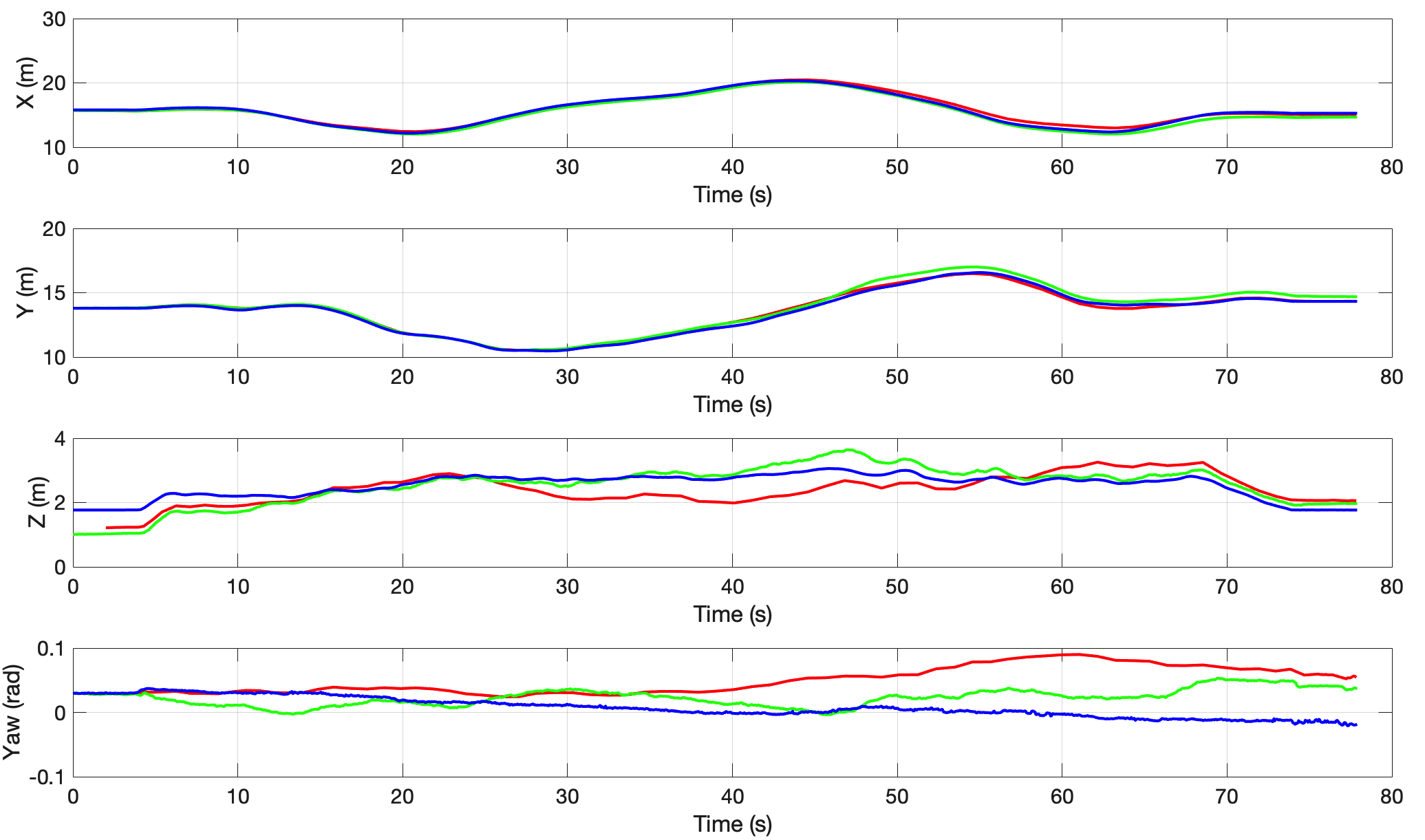}
    \includegraphics[width=0.47\textwidth]{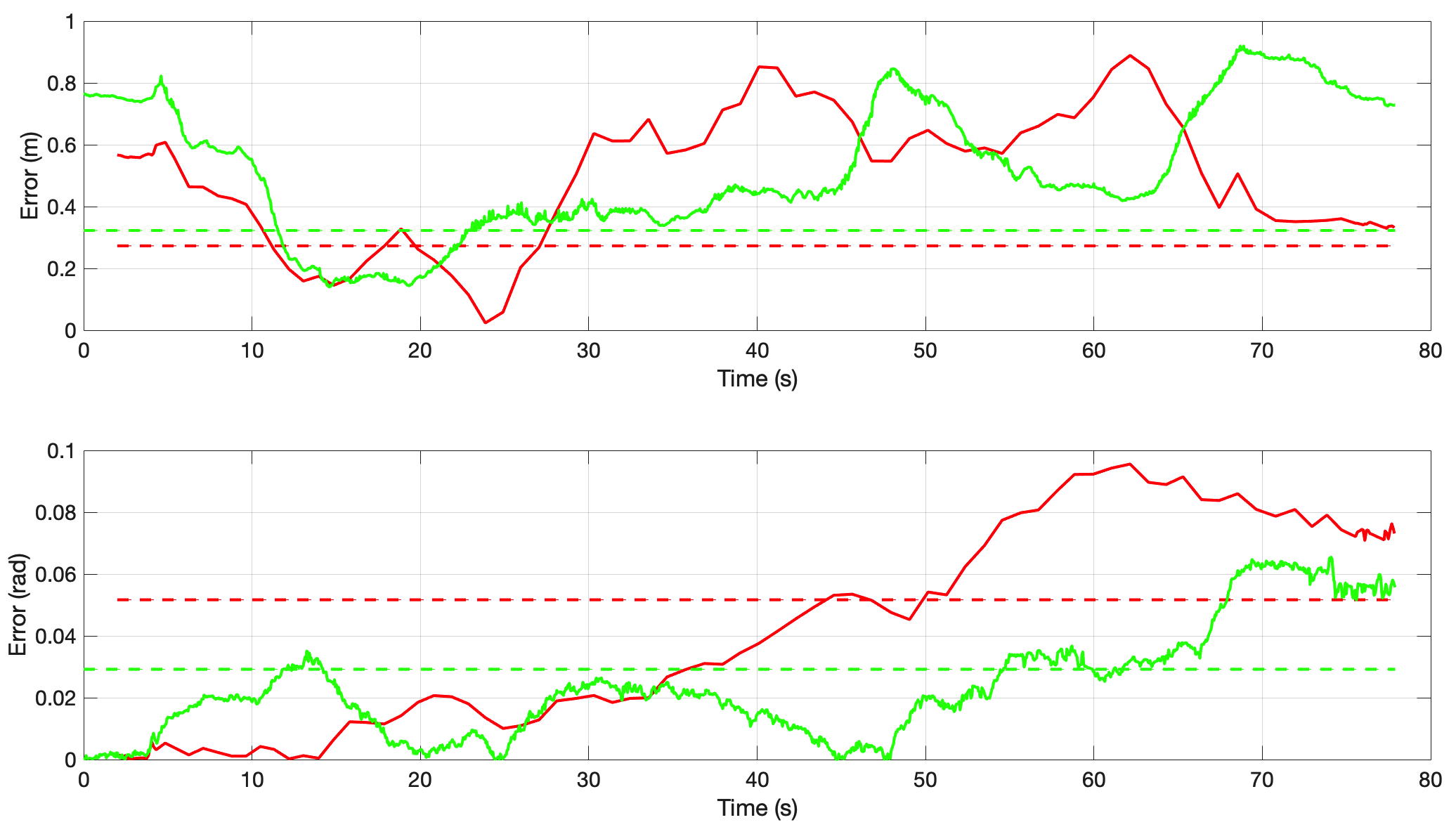}
    \caption{\revisionFer{Robot localization estimation and errors. The red line stands for the MCL3D estimation, the green line denotes the estimation of our method and blue line stands for the ground-truth robot position and orientation (Left) Estimated robot position in X, Y, Z and Yaw angle. (Right) Computed instant error in position (meters) and yaw angle (radians) with respect to the ground-truth. Dashed lines shows the corresponding RMSE error.} }
    \label{fig:rosin}
\end{figure}

\revisionFer{Figure \ref{fig:rosin} shows the resulted robot position and orientation estimations, and also the computed errors with respect to the ground-truth. In this experiment only LIDAR information and the environment map was used, no other sensor were used. Both algorithms have been setup with the same general parameters and make use of $1000$ particles. The LIDAR odometry presented in Section \ref{sec:3d_mapping} was used to compute the robot odometry. It can be seen how the errors in position are approximately the same in both approaches. Both approaches follow the ground truth position with small errors in X and Y, while most of the errors are introduced in the Z estimation. On the other hand, the errors in the yaw estimation are clearly smaller in our approach, the RMSE error is roughly a $40\%$ smaller. This improvement mostly comes from the new particle prediction model presented in (\ref{eq:prediction}). The possibility to introduce small prediction noise in the particle's evolution helps to model small inaccuracies coming for the odometric system. }

\revisionFer{As previously commented, the errors in Z and yaw use to be predominant in our approach. This is related with the fact the 3D LIDAR mounted in the aerial robot mostly perceives the walls and structures, but not the floor. On the other hand, the yaw angle is also prone to drift due to error accumulation in the odometry. Section \ref{sec:loc} proposes the integration of the localization approach with other sensors such as compass, GPS and altimeter to reduce the increase the accuracy and robustness of the system. Thus, Table \ref{tab:rmse_loc} presents the computed RMSE error in the estimated robot's position and orientation for different combinations of sensors.} 

\begin{table}[h]
\begin{center}
\begin{tabular}{| l | c | c |}
\hline
Approach & RMSE (m) & RMSE (rad) \\ 
\hline
MCL3D & 0.2729 & 0.0517 \\
Ours & 0.3226 & 0.0292 \\
Ours+Alt & 0.2736 & 0.0245 \\ 
Ours+Yaw & 0.2555 & 0.0015 \\
Ours+Alt+Yaw & 0.2167 & 0.0016 \\
Ours+Gps & 0.3428 & 0.0302 \\
Ours+Gps+Alt & 0.2777 & 0.0241 \\
Ours+Gps+Alt+Yaw & 0.2205 & 0.0016 \\
Ours+Gps+Yaw & 0.2540 & 0.0016 \\
\hline
\end{tabular}
\caption{Estimated RMSE localization error of the proposed approach}
\label{tab:rmse_loc}
\end{center}
\end{table}

\revisionFer{The sensors used in Table \ref{tab:rmse_loc} have been simulated based on the ground truth. Given the truth position, altitude and yaw angle, noisy readings of each sensor have been computed. We injected Gaussian errors in the simulated measurements in order to mimic the behaviour of a real sensor. It can be seen how the introduction of the yaw angle has a significant impact in the general estimation, not only in robot yaw (which is is obvious) but also in position, as can be seen in the implementation using yaw. On the other hand, the altimeter also benefits the robot position estimation. However, GPS does not provide a clear improvement in the estimation. This is probably produced by the relatively high sensor noise (0.5 m in X and Y position), which is in the order of the errors computed based in the map/point-cloud matching. However, having alternative estimation of robot position is beneficial from the point of view of redundancy, reason why we performed experiments also with GPS integration.}

\begin{figure}[!t]
    \centering
    \includegraphics[width=0.98\textwidth]{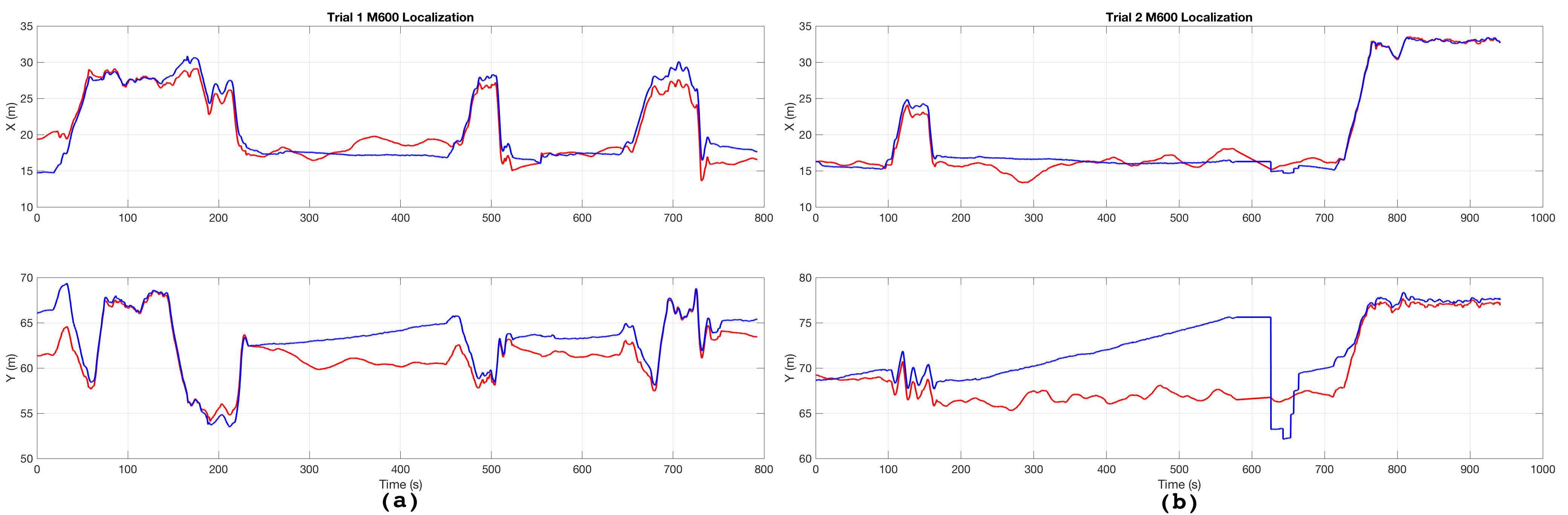}
    \caption{DJI M600 X-Y localization at the MBZIRC arena during execution of Trials 1 and 2. (a) Estimated XY position in Trial 1. (b) Estimated XY position in Trial 2.  Red line denotes the MCL estimation and blue line denotes the GPS estimation.}
    \label{fig:trials_m600_loc}
\end{figure}

\revisionFer{The proposed approach integrating compass, altimeter and GPS was used and tested in experiments involving real robot operations during the MBZIRC trials.} Figure \ref{fig:trials_m600_loc} shows the estimated localization of our DJI M600 during two experiments in the MBZIRC arena. The figure also shows the estimated GPS position of the DJI M600 during the experiments (blue line in the graph). We can see GPS estimation issues in both experiments. Thus, in Fig. \ref{fig:trials_m600_loc}-b it can be seen how from second 175 to 625 approximately, the GPS estimation evolves at constant velocity in Y axis, with a 15 meters jump in Y at second 625. During this period the drone was landed so the robot position must stay approximately constant. 

\revisionFer{Figure \ref{fig:trials_m600_loc}-a also shows} constant velocity integration in axis Y from second 225 to 475 approximately. As previously noted, the GPS eventually recovers the right estimation, converging to the MCL estimation, which stay stable during the \revisionDat{whole} experiment. 

\revisionLuis{These GPS errors presented in Fig. \ref{fig:trials_m600_loc} were common during the MBZIRC 2020 execution. This behaviour was detected during the rehearsal days, and we finally decided to set the value of $w^{gps}$ from (\ref{eq:alpha}) to $0$ to avoid integrating outliers into the estimation, leading to a complete GPS-free localization approach}.

%% file: 04_Navigation.tex

\revisionLuis{The challenge requires the robots to reach arbitrary free poses in the map (outside and inside the building) to look for fires and to attack them once detected.}
\revisionFer{A two-stage autonomous navigation system (global and local planners) has been implemented in order to deal with this environment.}
The Global Planner computes a global path on the map frame. The inputs are the goal to be reached, the static global map of the environment and the current robot localization given by the \revisionFer{localization} approach described in the previous section. 
The output is a feasible and safe path that should be followed in order to reach the goal. 
The Local Planner is in charge of following the global path and dealing with dynamic and unmapped obstacles detected by the onboard 3D LIDAR. 
It takes as input the global path 
and a local map built by using the measurements of the onboard sensors. 
As output, it provides a free-collision local trajectory to the path tracker module. \revisionLuis{In the following, we describe the main techniques for each module}.

\revisionLuis{\subsection{Path planning}}

\revisionLuis{The application considered does not  require precise dynamic motion planning for the vehicle. The main consideration is to provide safe  paths within the workspace, handling the vehicle dynamics in the path tracker. At the same time,} we need fast planning capabilities, as in the case of the local planner a new local plan should be generated whenever a new LIDAR measure is obtained (and therefore a new local map is generated), at an approximate rate of 10Hz. 

\revisionJac{
Different types of approaches for tackling 3D path planning have been proposed in the literature: discrete search optimal algorithms, sampling-based algorithms and bioinspired algorithms. Related to the first type, the A* algorithm has been used in several works \cite{Filippis2012,Sanchez2019,Zhan2014}. Another algorithm is the Lazy Theta* algorithm, which is based on the A* algorithm \cite{THETA}. The Lazy Theta* algorithm, an any-angle find-path algorithm, computes shorter paths than the A* algorithm because it does not constrain paths to be formed by graph edges (to find short “any-angle” paths). Sampling-based algorithms such as RRT (Rapidly exploring Random Tree) have also been used in path planning \cite{Lin2017}. An comparison between A* algorithm and RRT algorithm is done in \cite{Zammit2020}, both algorithms can be applied in real–time. The advantage of the graph-based algorithms with respect to sampling-based algorithms is that the computed path is closer to the shortest path, and the solution is deterministic. Bioinspired approaches imitate the behavior of humans or other animals, it is worth to mention Particle Swam Optimization (PSO) \cite{Alejo2014}, and also genetic algorithms \cite{Conde2012}. These algorithms generally require significant computational resources, and their performance may vary depending on the scenario considered.}
\revisionJac{Considering the Challenge 3, we need to implement a safe and precise waypoint navigation system which rapidly plan paths. The Lazy Theta* algorithm presents the most suitable characteristics and is chosen with respect to other planning algorithms due to its computational load and its high repeteability. 
}


Thus, \revisionLuis{the path} planners are based on the Lazy Theta* algorithm \cite{THETA}, a fast and reliable heuristic planner. \revisionJac{However, we have modified the original Lazy Theta* algorithm to foster safety while still maintaining its good computation time. Algorithm \ref{l:lazy} shows the pseudo-code of the original algorithm and highlights the two modifications implemented (in red).} 


\begin{algorithm}[tb!]
 \SetKwProg{UpdateVertex}{UpdateVertex(s,s')}{}{end}
 \SetKwProg{ComputeCost}{ComputeCost(s,s')}{}{end}
 \SetKwProg{SetVertex}{SetVertex(s)}{}{end}
 \SetKwProg{IF}{If}{ then}{}
 \SetKwProg{FOREACH}{foreach}{ do}{}
 \SetKwProg{WHILE}{while}{ do}{}
 \SetKwProg{main}{Main()}{}{end}
  \main{}{
    open := closed := $\emptyset$\;
    g($s_{start}$) := 0\;
    parent($s_{start}$):=$s_{start}$\;
    \tcc{$g(s)$ is the length of the shortest path from $s_{start}$ to $s$}
    \tcc{$h(s)$ is the approximated distance from the goal to the vertex}
    open.Insert($s_{start}$,$g(s_{start})+h(s_{start})$)\;
    \WHILE{open $\neq \emptyset$}{
    s := open.Pop()\;
    SetVertex(s)\;
        \IF{ $s = s_{goal}$ }{
            \Return{"path found"}\;
        }
        closed := closed $\cup s$\;
        \FOREACH{$s' \in nghbr_{vis}(s)$ }{
            \IF{$s' \not\in$ closed }{
                \IF{$s' \not\in$ open }{
                g($s'$) :=  $\infty$\;
                parent($s'$) := NULL\;
            }
            UpdateVertex($s$,$s'$)\;
            }
        }
        \Return{"no path found"}\;
     }
  }
  
  \UpdateVertex{}{
  $g_{old}$ := $g(s')$\;
  ComputeCost(s,s')\;
  
  \IF{$g(s') < g_{old}$}{ 
    \IF{$s' \in $ open}{ 
        open.Remove(s')\;
      }
      open.Insert($s'$,$g(s')+h(s')$)\;
    }
  }
  \ComputeCost{}{
    \tcc{Path 2}
    \tcc{$c$ is the cost to reach a node}
    \IF{$g(parent(s)) + c(parent(s),s') + \textcolor{red}{C_w \cdot dist\_obst} < g(s')$ }{
        parent($s'$) := parent(s)\;
        $g(s')$ := $g(parent(s)) + c(parent(s),s') + \textcolor{red}{C_w \cdot dist\_obst}$ \;
        }
  }
  \SetVertex{}{
    \IF{NOT lineofsight(parent(s),s) \& \textcolor{red}{lineofsight(parent(s),s) $<$ max}}{
        \tcc{Path 1}
        parent(s) := \it{argmin} ($s' \in nghbr_{vis} \cap closed(g(s')+c(s,s'))$)\;
        g(s) := \it{min} ($s' \in nghbr_{vis} \cap closed(g(s')+c(s,s'))$)\;
        }
    }
\caption{Lazy Theta* with the introduced modifications in red.}
\label{l:lazy}
\end{algorithm}


\revisionLuis{These two modifications of} the Lazy Theta* algorithm have the following effects: 

\begin{itemize}
    \item \textbf{Cost to reach a node}. We added a cost component to the Euclidean distance in the original algorithm \revisionJac{(see lines 33 and 35 of the Algorithm \ref{l:lazy})}. For each node in the grid, this cost is the distance to the nearest obstacle \revisionJac{(\texttt{dist\_obst} in lines 33 and 35)}. Then, the cost to reach a node is now a sum of the \revisionLuis{path length} 
    and the new cost component for the nodes that define the trajectory. The latter component is weighted by a factor $C_w$ between 0 and 1. 
    This results in paths with larger safe margins with respect to obstacles, depending on the value of the weight factor.
    \item \textbf{Line of sight}. The line of sight is limited in order to allow the algorithm to link two nodes only if they are not further than a given distance. Otherwise, the first modification would have no effect because the original Lazy-Theta* algorithm joins nodes if there is a direct line of sight between them, regardless of the cost. \revisionJac{Thus, a maximum line of sight distance is included, \texttt{max} in line 38 of the Algorithm \ref{l:lazy}, allowing the algorithm to link two nodes only if they are not further than this distance.} \revisionLuis{If this distance is set to $\infty$ then we recover the original Lazy-Theta* algorithm. 
    This will affect directly the number of expanded nodes.}
\end{itemize}

\begin{figure}[!t]
	\centering
    \includegraphics[width = 1\textwidth]{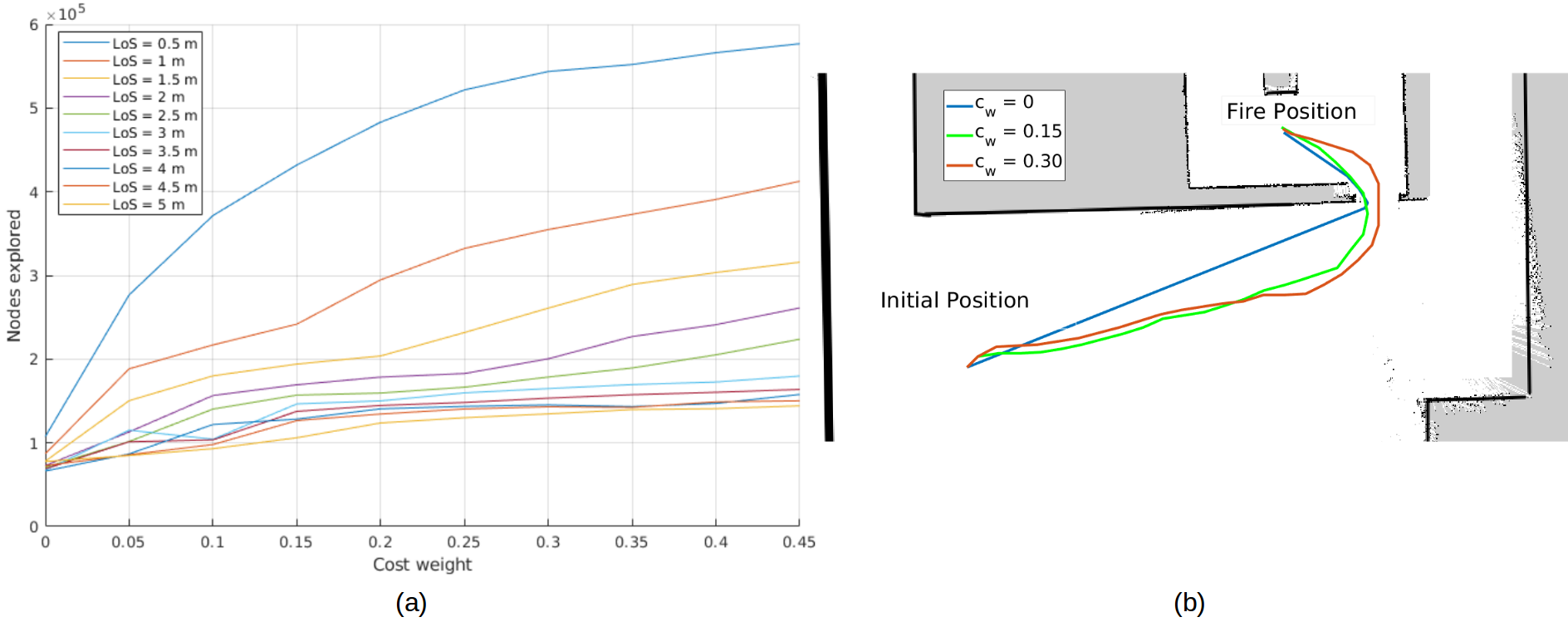}
	\caption{\revisionJac{(a): Nodes explored depending on the weight factor and the line of sight.} \revisionJac{(b): 2D trajectories computed for the UGV coming into the building (2D map of navigation) by considering different values of the weight factor, $C_w$.}}
	\label{fig:planners}
\end{figure}

These two modifications \revisionJac{should} provide paths with larger safety margins with respect to obstacles at the cost of exploring more nodes. The quality and the computation time of these paths \revisionJac{will} depend on the weight factor and maximum line of sight distance, so the values of these parameters \revisionJac{in both planners} \revisionJac{should be tuned}.

\subsubsection{UGV planner}

\revisionJac{The influence of the two modifications of the Lazy Theta* algorithm has been analyzed in the 2D planner of the UGV. Figure \ref{fig:planners} presents how the weight factor, $C_w$, and line of sight, $LoS$, influence the trajectories computed by the 2D planner. Figure \ref{fig:planners}-(a) shows the number of explored nodes with different weight factors and maximum line of sight distances. The number of explored nodes (and thus, the computation time) augments as the weight factor increases for every value of the line of sight. This increase takes also place as the line of sight decreases.} Figure \ref{fig:planners}-\revisionJac{b} shows trajectories computed with different weight factors and a maximum line of sight distance of 1.5m. The weight factors 0.15 and 0.3 generate a safer trajectory than the one obtained with the original algorithm (blue line). This can be appreciated when robot crosses the door. 

\revisionJac{In order to properly select the right values of $C_{w}$ and $LoS$, we have to consider that the local planner must generate trajectories at an approximate rate of 10Hz to ensure fast planning capabilities. Additionally, we must choose a good compromise solution between safety and trajectory length. The chosen values are $LoS=1.5$ and $C_{w}=0.15$. It can be seen in Fig. \ref{fig:planners}-(a) that the computation time considering $C_{w}=0.15$ is smaller than considering $C_{w}=0.3$, while giving similar results. Although the number of explored nodes doubles those of the original algorithm, it still ensures the rate for fast planning capabilities.}





\subsubsection{UAV planner}

\begin{figure}[!t]
	\centering
    \includegraphics[width = 1\textwidth]{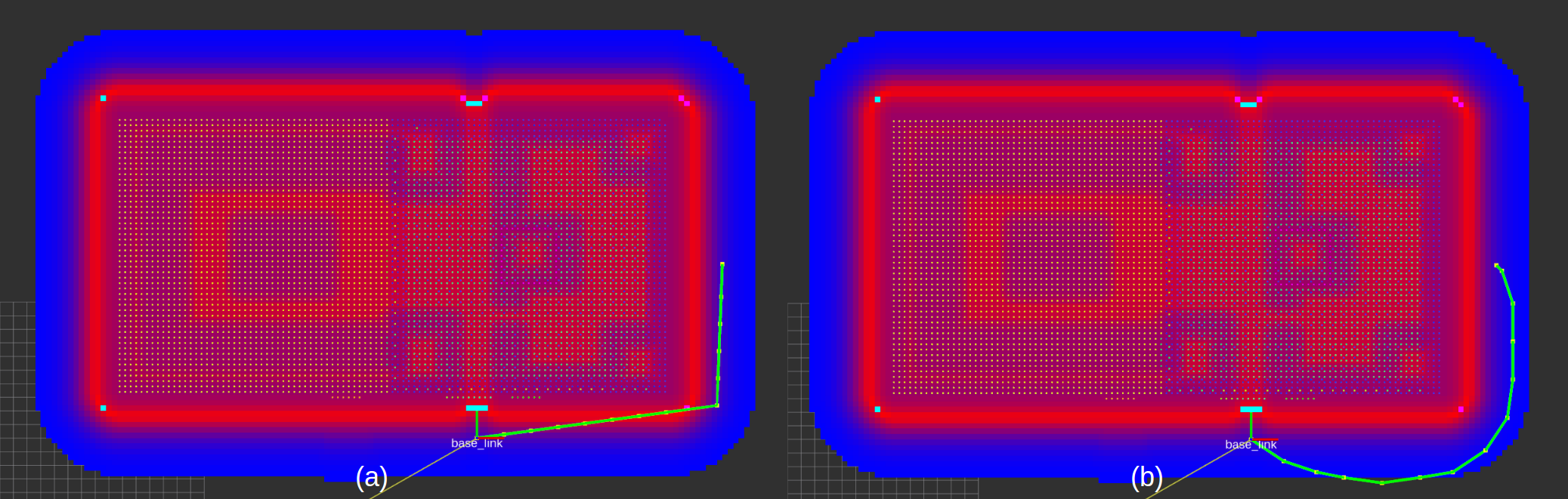}
	\caption{\revisionJac{Influence of the line of sight and weight factor on the safety margin (top view). The distance cost is also shown, red area around the building is the area with a higher distance cost. (a) Trajectory computed without line of sight and the weight factor 0 (original Lazy-Theta*), (b) Trajectory computed considering that line of sight is 2 and the weight factor 0.3. The base\_link frame is the initial position.}}
	\label{fig:3Dplanner_safety}
\end{figure}

\begin{figure}[!t]
	\centering
    \includegraphics[width = 1\textwidth]{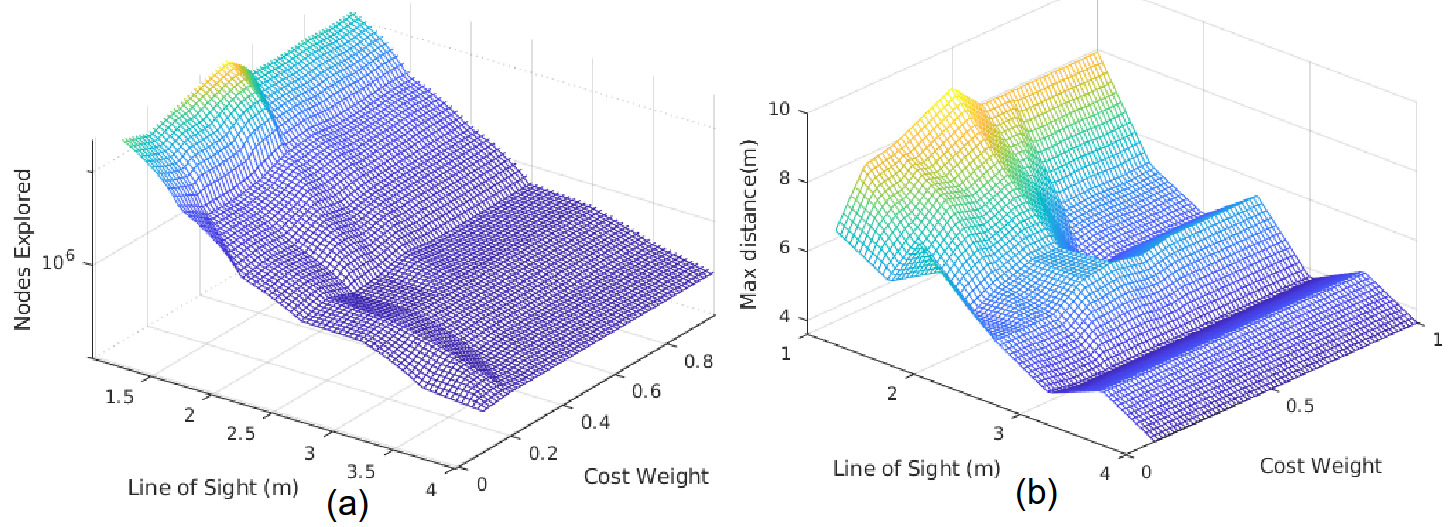}
	\caption{\revisionJac{(a) Nodes explored to compute the trajectory with lines of sight from 1.0 to 4.0 and weight factors between 0 and 1 for each line of sight, (b) Influence of the line of sight and weight factor on the safety margin. The trajectories considered are the computed in the scenario shown in Figure \ref{fig:3Dplanner_safety} with the same initial point and goal.}}
	\label{fig:3Dplanner_nodes_safety}
\end{figure}

\revisionJac{The same analysis has been performed in the 3D planner of the UAV. 
Figure \ref{fig:3Dplanner_safety} illustrates how the safety is fostered by showing a trajectory computed with the original Lazy Theta* algorithm and another one considering the modified algorithm with $LoS=2$ and $C_{w}=0.3$. The trajectory length in the first one is 14m and in the second one is 15.6m. This larger trajectory makes the safety margin with respect to the building to increase during the flight. The closest distance between the building and the path is 0.5m with the original algorithm. However, the trajectory of the modified algorithm departs from the building (up to 5.1m away) and the minimum distance, 2.5m, takes place when the UAV reaches the goal.}

\revisionJac{Figure \ref{fig:3Dplanner_nodes_safety}(a) shows the explored nodes with different values of the line of sight and weight factors. 
The explored nodes decrease as the line of sight increases. Instead, the number of nodes fluctuates a little with different values of the weight factor. Considering the original algorithm, $LoS$ is not limited and $C_{w}=0$, the number of explored nodes is much less, 28,851. Therefore, it is noteworthy how the modified algorithm influences the number of explored nodes.}

\revisionJac{Figure \ref{fig:3Dplanner_nodes_safety}(b) shows the maximum distance the paths moves away from the building considering different values of the line of sight and weight factor. In the case of the path computed with the original algorithm, this distance is always smaller than 2.5m (see Fig. \ref{fig:3Dplanner_safety}-a). Instead, the UAV is always at least 3.5m away as the modified algorithm is used, and even this distance is greater than 4m as $LoS<3$. The highest distances are achieved with $LoS=1$. Regarding the closest the path gets to the building, it is never shorter than 2.5m (distance between the goal and the building) as $LoS<2.5$. This distance is 2.4m  for $LoS=2.5$ and between 1.7m and 2.4m in the rest of cases, $LoS=3.0, 3.5, 4$. Therefore, safer and smoother trajectories are computed thanks to the modifications performed. An example is the path shown in Fig. \ref{fig:3Dplanner_safety}-b.}


\revisionJac{Finally, we decided not to enter the building with UAVs in the trials. Although the advantages of the modified planning algorithm have been shown and the compromise between safety and planning capabilities can be reached, this decision guaranteed safety 
because the closest distance to the building would be the distance of the inspection points to the facade. For this reason, in the competition we did not limit the line of sight neither consider the distance cost, $C_{w}=0$, in the 3D planner. This allowed us to maintain the fast planning capabilities of the 3D planner.}

\subsection{Path tracker}

The Path Tracker module computes the velocity commands to navigate and follow the trajectory computed by the Local Planner in both platforms. \revisionDat{The UGV makes use of a pure pursuit (PP) \cite{Kelly-1994-13704,park_14} controller that adapts the angular velocity of the robot to reach the closest point into the commanded trajectory. The main idea of PP is to calculate a real-time target waypoint (WP) which lays in the target path and is situated at a look-ahead (L) distance from the vehicle.  When applied to non-holonomic robots, PP generates a commanded heading change $\Delta\theta=arctg(y/x)$ that is necessary to make the robot point towards WP, where $x$ and $y$ are the coordinates of WP in the reference frame of the robot.}



\begin{figure}[!t]
	\centering
     \includegraphics[width=0.45\textwidth]{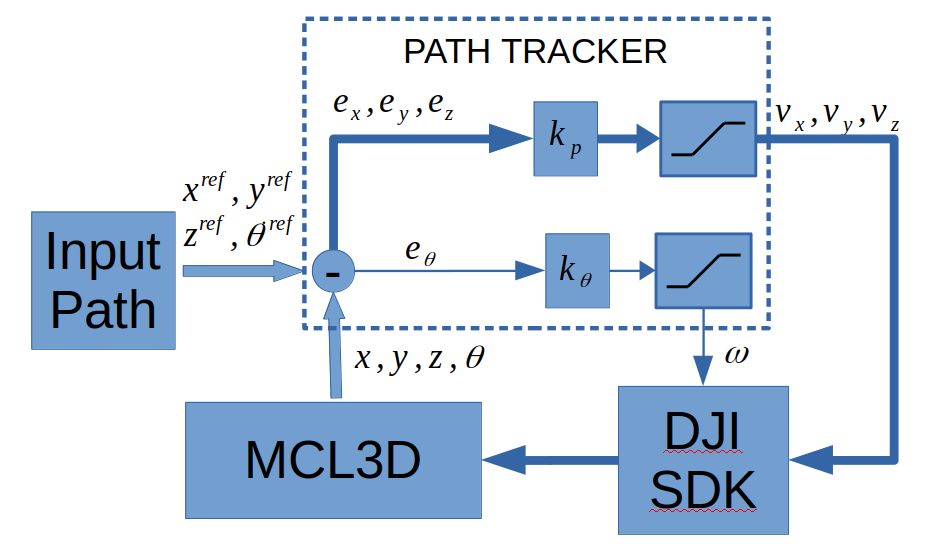}
	\caption{(a) Basic control scheme of the path tracker implemented on the UAVs.}
	\label{fig:tracker}
\end{figure}

On the other hand, the UAVs make use of the command velocity interface provided by the DJI software. This interface allows commanding linear velocities in X, Y and Z axes, and angular rates in yaw. \revisionDat{A saturated proportional controller (see Fig. \ref{fig:tracker}) has been implemented to follow the commanded local trajectory in the UAVs. This control scheme gives us a trapezoidal velocity profile in each axis in order to reach the current destination waypoint while ensuring convergence to the final destination as we have a first order system.}

\subsection{Experiments}

\begin{figure}[!t]
	\centering
     \includegraphics[width=0.9\textwidth]{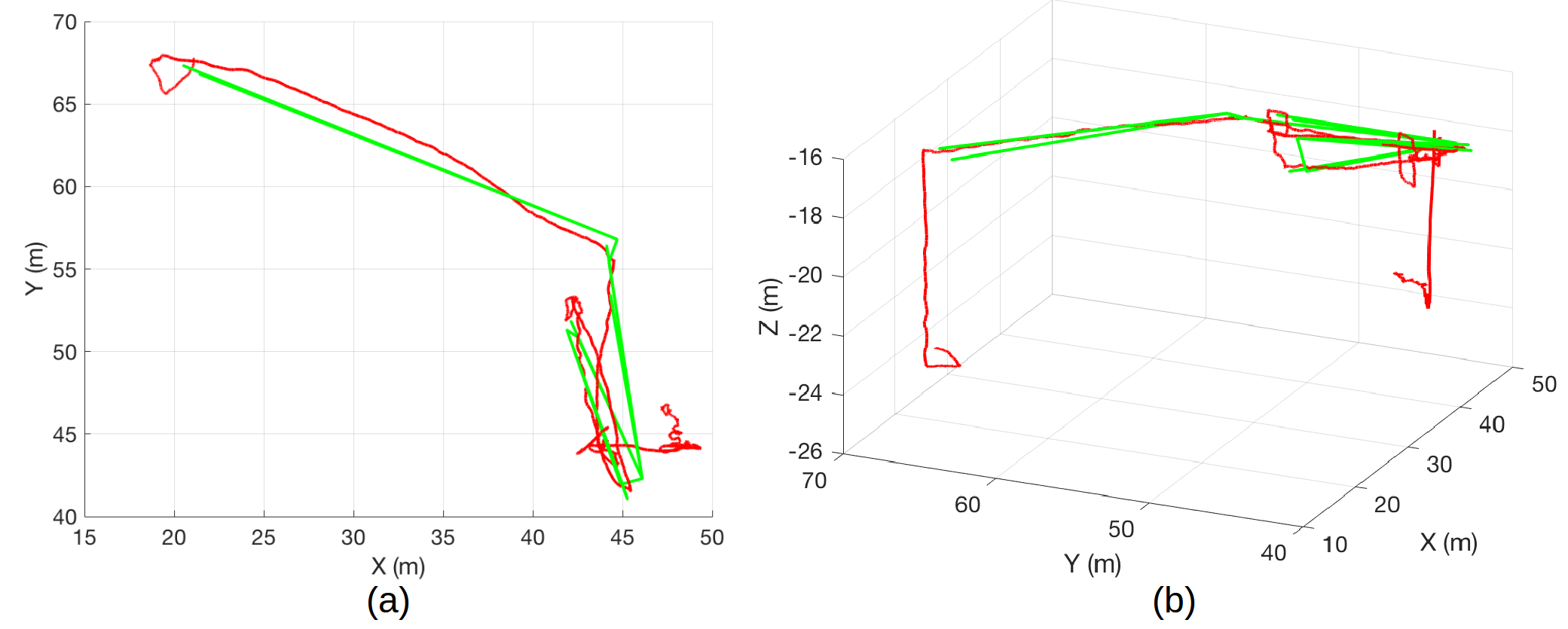}
	\caption{Top view (a) and 3D view (b) of the global planned trajectory (green line) and its associated executed trajectory, as estimated by our localization system (red line) during a flight of the DJI Matrice M210 in the second trial for Challenge 3.}
	\label{fig:planner_uav}
\end{figure}

Figure \ref{fig:planner_uav} presents the trajectory followed by the M210 UAV in one experiment, in red, alongside with the global planned trajectory in green. As we use a local planner in order to check for unmodeled obstacles on the environment, the tracking of the trajectory on the horizontal plane is not perfect and it tends to take shortcuts on the corners whenever possible. On the other hand, the system follows the altitude reference more closely.

The 3D planner obtained a remarkable performance that allowed us to perform all experiments during the MBZIRC competition without any collisions. The computation times of the global plan is of less than a second in its onboard NUC7i7 and using a single core. On the other hand, the local planner computes trajectories at 7Hz approximately \revisionJac{which ensures fast planning capabilities of the 3D local planner}.

%% file: 05_FireExtinguishing.tex
Our approach for the fire extinguishing task in the MBZIRC - Challenge 3 consists of two modules: the perception module detects fires and estimates their positions with a thermal camera; and an extinguishing module that attacks the fire by dropping a blanket or ejecting water.

\subsection{Fire perception}\label{sec:firedetect}

The module of fire perception detects fires within the scene using as input the data from a thermal camera. As output, it indicates if a fire has been detected, as well as its 2D coordinates on the image plane and 3D coordinates in the scenario.  

All fires in the competition were simulated using a thermal heated plate that reached temperatures up to 110º. The fires on the facades featured additionally a real fire using gas, while the remaining fires were also visually simulated by using a flame simulator using a red silk.

In our solution, we aimed for a fully infrared-based detection of the heat source. The main sensor related to fire perception is the miniature infrared thermal camera from Seek Thermal mentioned above. The camera can detect temperatures between -40º to 330º Celsius. We forked a ROS driver from ETHZ and made additional modifications\footnote{\url{https://github.com/robotics-upo/seekthermal_ros}}. The driver provides a grayscale temperature image, as well as a radiometric map.

\begin{figure}[!t]
	\centering
    \includegraphics[width=1.0\textwidth]{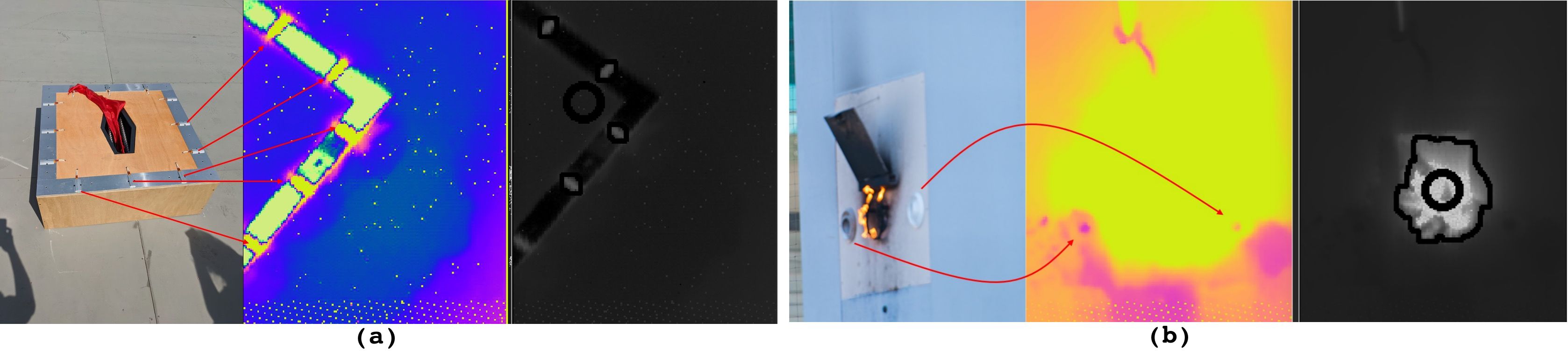}
	\caption{(a) Simulated ground fire, colored temperature image and segmentation using the radiometric values (the circle shows the mean position of the fire). (b) Simulated facade fire, colored temperature image and segmentation using the radiometric values. The original temperature image gets saturated, while working with the radiometric field allows for a detailed segmentation. }
	\label{fig:firedetection}
\end{figure}

Fire detection is carried out by analyzing the images provided by the infrared cameras. A calibration procedure transforms the raw radiometric data obtained by the cameras to temperature ranges. While obtaining absolute temperature values from radiometry is a very complex process, dependant on the material emissivity (for instance, a metal can radiate less IR energy for the same temperature than other materials with larger emissivity), atmospheric attenuation, and many other factors, here we are more interested on clear relative temperature differences, which simplifies the calibration procedure. The calibration is performed for the different kind of fires once. Then, given the clear difference of temperatures between the fire and the surroundings, a thresholding operation is performed to segment the fire pixels. As a result, the fire detection module provides coordinates on the image plane of fire spots. Figure \ref{fig:firedetection} shows detection results obtained during the competition. \revisionLuis{During the competition, no false positives were generated by using the method.}

\subsection{Fire position estimation}\label{sec:fireposition}

The fire segmentation results are used for the visual servoing final approach to extinguish the fire (see Section \ref{subsec:extinguish}). However, during the exploration of the scenario we need to estimate the 3D position of the fire spots to plan the approximation goals for fire extinguishing.

From the fire position in pixel coordinates it is not possible to the estimate the 3D position of the fires, as there is no information about the range to the fires. In order to obtain the range, the LIDAR information is considered. The LIDAR values are mapped to the image coordinates given the known static transformation between sensors and camera internal calibration. This way they can be associated to the fire object, so that a range can be estimated. This is not always possible given the resolution of the LIDAR (and depending on the actual distance to the fire). If not range information is available, the range is estimated considering the maximum distances according to the 3D map of the environment. For the fires located on the floor, the height of the UAV is considered to estimate the position of the fire. 

Each \revisionLuis{3D} measurement is associated with a covariance matrix that reflects the uncertainty on the range, which is adapted according to the estimated range (when no range is available, a large uncertainty is associated to it), and other aspects like the position of the robot. Then, an Information Filter is used to fuse in time the measurements about the position of the fire from different view points in order to triangulate the fire position, as in our previous work \cite{triangicra15}.

\subsection{Fire extinguishing}\label{sec:fireextinguish}
\label{subsec:extinguish}

This module is in charge of confirming the fire presence, aiming adequately the extinguisher system, and releasing the water or the blanket. The module assumes that the robot is already located at an attacking distance to the fire. Three tasks should be accomplished separately in each robot: confirming, centering and extinguishing.

\revisionLuis{\subsubsection{Fire extinguishing with UGV using water}}\label{sec:fireextinguishugv}

The \revisionLuis{confirming phase is done by scanning} with the Pan\&Tilt unit onboard the UGV to look for the precise position of the fire. To this end, we command a predefined trajectory to the pan and tilt unit using position commands. At this point, the Fire Detection module is activated. 

Whenever the Fire Detection module confirms a fire, \revisionLuis{the centering phase use a velocity visual servoing proportional controller to command the Pan\&Tilt unit joints. The error signal is the pixel difference between the detected fire position on the image plane and the center of the locked-on target zone}. We have empirically defined this zone in such a way that the water ejecting mechanism hits on target. This zone is indicated in an area of the image and it depends on both the height and distance to the target. 

Once the extinguisher has locked on the fire, \revisionLuis{the extinguishing phase begins, and the system ejects water}. As a final note, it is difficult to perceive the trajectory of the water with the onboard sensors. Thus, to increase the chances of successfully hitting the target, the pan\&tilt unit performs a cross-shaped movement while ejecting water during 60 seconds.   

\revisionLuis{\subsubsection{Fire extinguishing with UAV using water}}\label{sec:firefacadewater}

\begin{figure}[!t]
	\centering
	\includegraphics[width = 1.0\textwidth]{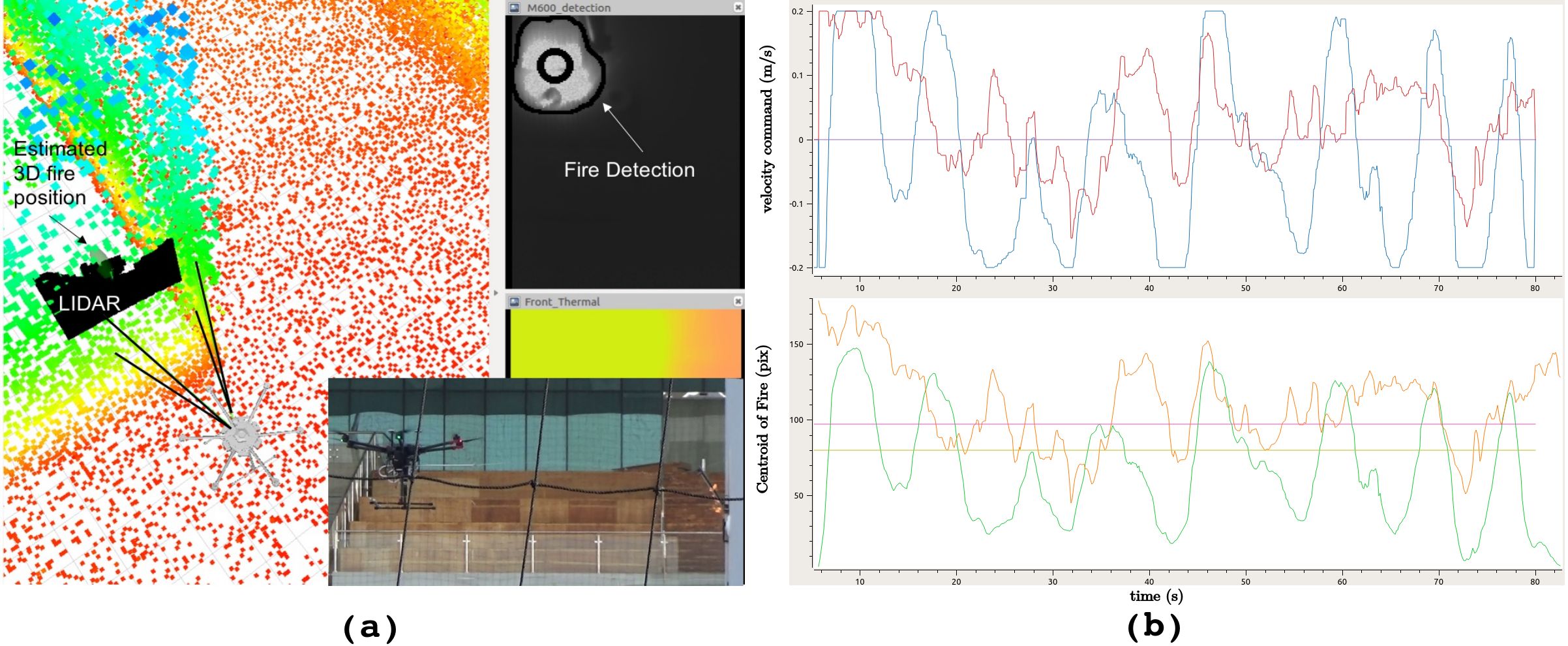}
	\caption{Facade fire extinction with an UAV using water. (a) The thermal camera is used to detect the fire. The 3D position of the fire is estimated by using the LIDAR information and tracked using an Information Filter. (b) Visual servoing for extinguishing. Top: commanded UAV speeds generated by the control system on $y$ (horizontal UAV body axis, parallel to the wall, blue) and $z$ (vertical UAV body axis, red). Bottom: $u$ (columns, green) and $v$ (rows, orange) image coordinates of the centroid of the fire detected on the image. The $u$ and $v$ target coordinates are plotted in pink and light green, respectively.}
	\label{fig:control_facade_m600}
\end{figure}

To extinguish the fires on the facade, once the UAV is positioned in the attack position, the UAV first performs a manoeuvre to confirm that there is a fire in the area (to compensate for imprecision on reaching the fire attacking waypoint or on the estimated 3D fire position). This manoeuvre consists of performing a square motion parallel to the facade of the building. 

As soon as a fire is detected the square sequence stops. We use a proportional velocity controller in the local Y and Z axes of the UAV so that the detected position of the fire on the image plane enters into the locked on target zone. As in \revisionLuis{the UGV case}, we have empirically defined the locked on target area on the image assuming a distance of attack to the fire. Once the fire is on the locked on target zone, the water ejecting starts. As the UAV can be subjected to disturbances while flying which may cause the UAV move outside the lockup zone, the velocity controller keeps regulating the position during the ejection.

Figure \ref{fig:control_facade_m600} presents an example from the competition (during the Grand Challenge). It shows the output of our control system as a function of the coordinates of the centroid of the detected fire region on the thermal image. This control system sends velocity commands to the M600 autopilot in order to center the fire in the image.  This particular fire was subjected to lateral wind gust of up to 8 m/s. It can be seen in Fig. \ref{fig:control_facade_m600} how this provoked oscillations in the control in the lateral velocity, making it difficult to adequately point the water hose. The fire was never completely lost during the water ejection and the fire was considered partially extinguished.

For the fires inside the building, as before, first the drone starts a fire confirmation sequence. In this case, to prioritize the safety of the platform, the maneuver consists only of yaw rotations in place. If a heat trace is found, then a proportional velocity controller in yaw is executed to center the fire horizontally on the image plane. When this is achieved the water ejecting starts, keeping the controller for correcting disturbances.

\revisionLuis{\subsubsection{Fire extinguishing with UAV using blankets}}\label{sec:fireextblanket}


\revisionLuis{In this case} the UAV performs a confirming fire manoeuvre similar to the one done for facade wall fires, but this time the UAV moves in a plane parallel to the ground. As the blanket must cover the fire, the extinguishing action is more convoluted. Once the fire is confirmed on the image, the UAV is placed just above the fire. Then the drone descends to be more accurate when dropping the blanket. Afterwards, the UAV moves backwards until the fire is out of image and stop. Just in that moment, two electromagnets from the extinguishing system switch off to unroll the blanket.
From there, the UAV moves forward passing throw the fire, and so detecting the fire again with the thermal camera. The UAV will continue this movement, over passing the fire, and just when the thermal camera stop detecting the fire, the last four magnets will switch off and so the blanket drops covering the fire. From there, a fast move upward is performed in order to avoid any collision with the deployed blankets.

%% file: 06_ExecutiveLayer.tex
\revisionLuis{\subsection{System architecture and atomic tasks}}

\begin{figure}[!t]
	\centering
    \includegraphics[width = 0.7\textwidth]{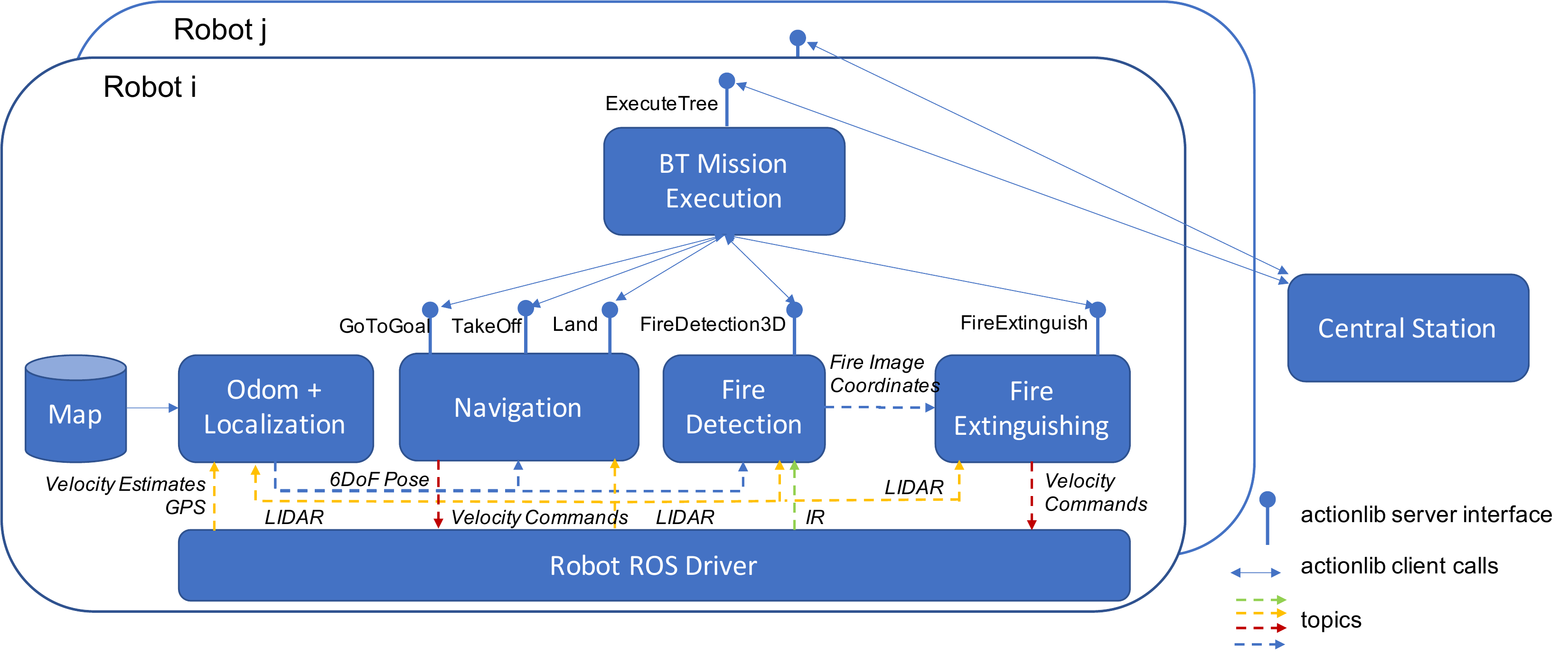}
	\caption{A depiction of the architecture. Each robot has operational autonomy to execute missions, and all processing takes place onboard, where all the enclosed modules enclosed run. A central Control Station can order the robots to execute missions through the BT Mission Execution node.}
	\label{fig:actionlib}
\end{figure}

All former components need to be integrated and work jointly to achieve the missions of the Challenge. This integration relies on the ROS framework, using the \texttt{ros-kinetic} distribution under Ubuntu 16.04. The algorithms described above have been implemented as ROS nodes. 

The basic architecture of the system is shown in Fig. \ref{fig:actionlib}. Continuous data flows between nodes are implemented through ROS topics. 
Some of these nodes also provide atomic tasks that will be composed to create the whole missions. \revisionLuis{As abstraction} for such tasks, we employ the ROS' \texttt{actionlib} package\footnote{\url{http://wiki.ros.org/actionlib}}, which offers a simple interface to preemptable tasks. In particular, we use the task model offered by the \texttt{SimpleActionServer} implementation (actions can be \texttt{ACTIVE} when running; \texttt{SUCCEEDED} when successfully finished; \texttt{CANCELED} if they cannot be processed; \texttt{ABORTED} if they cannot be completed; finally, actions can be also \texttt{PREEMPTED} by the client). The list of atomic tasks considered by each robot are shown in Table \ref{tab:actions}. These tasks are the same for all robots, but of course they have specialized implementations depending on the robot (for instance, \texttt{TakeOff} is not applicable to the UGV, and the \texttt{FireExtinguish} action has the three variants described in Section \ref{subsec:extinguish}).

\subsection{Mission definition and execution}

\begin{table}[!t]
\centering
 \begin{tabular}{||p{2.6cm} | c | p{7.5cm} |p{2.1cm}||} 
 \hline
 \texttt{actionlib} & Params & Final Status & Result \\ [0.5ex] 
 \hline\hline
 \texttt{TakeOff} & Height & \texttt{CANCELLED} if vehicle not ready \newline \texttt{SUCCEEDED} if height reached \newline \texttt{ABORTED} if height cannot be reached after takeoff    & None \\ 
 \hline
 \texttt{Land} & None & \texttt{CANCELLED} if vehicle already landed \newline \texttt{SUCCEEDED} landed \newline \texttt{ABORTED} if the UAV cannot land & None \\
 \hline
 \texttt{GoToGoal} \newline (Section \ref{sec:navigation}) & Waypoint (WP) & \texttt{CANCELLED} if no path to WP \newline  \texttt{SUCCEEDED} if WP reached \newline \texttt{ABORTED} if WP cannot be reached    & None \\
 \hline
 \texttt{FireDetection3D} \newline (Section \ref{sec:firedetect}) & Duration & \texttt{SUCCEEDED} if fire found within the given duration \newline \texttt{ABORTED} if fire not found & 3D position of fire \\
 \hline
 \texttt{FireExtinguish} \newline (Section \ref{sec:fireextinguish}) & None & \texttt{SUCCEEDED} if fire can be locked on \newline \texttt{ABORTED} if fire cannot be locked on  & None \\ [1ex] 
 \hline
\end{tabular}\label{tab:actions}
\caption{Atomic tasks that can be carried out by each robot. Actionlib interface, parameters required, potential final status, and final result reported.}
\end{table}

\revisionLuis{Given the atomic tasks that can be carried out by the modules described in the former sections,  we need a framework to define whole missions as plans that combine those tasks, and a mechanism to execute and supervise the defined task plans. This mechanism is the mission executive, which is a fundamental component to allow the robots to have operational autonomy \cite{gancet2007embedding,molina20executive}. A popular example of executive is the ROS SMACH system \cite{bohren2010smach}, which uses nested Finite State Machines (FSM) to represent and execute high-level plans. 
}


\revisionLuis{In our case}, we employ Behaviour Trees (BTs) \cite{TRO17Colledanchise} as the framework for mission definition and execution for the robots of the team. 
They are an alternative to FSMs that offer advantages in terms of modularity and reactivity. First originated as a tool for Non-Player Character (NPC) AI development in the video game industry, BTs are becoming widespread in robotics, mainly due to its modularity and simplicity \cite{TRO17Colledanchise,paxton2017,Colledanchise_2018}, including their use to define behaviours for UAVs \cite{ogren2012increasing,molina20executive}.

While a full description of the BT framework is out of the scope of this paper, here we summarize its main elements. BT model behaviors as hierarchical trees made up of nodes (an example can be seen in Fig. \ref{fig:bt-fireexploration}). Trees are traversed from top to bottom at a given tick rate following a set of well-defined rules and executing the tasks/commands associated to the nodes that are encountered while doing so. Nodes' statuses are reported back up in the chain and the flow changes accordingly. A status can be either \texttt{SUCCESS}, \texttt{FAILURE} or \texttt{RUNNING}. According to their functionality, nodes can be classified as:

\begin{itemize}
\item \textbf{Composite}: it controls the flow through the tree itself and are similar to control structures in structured programming languages. 
\item \textbf{Decorator}: it processes or modifies the status it receives from its child. 
\item \textbf{Leaf}: this is where the actual task is performed, the atomic tasks that the robot can carry out, or other functionalities. As such, these nodes cannot have any children.
\end{itemize}

As it can be seen from the classification above, BT decouple logic from actual tasks in a natural way. When developing a tree, one only should care about the leaf nodes. In this case, these leaves correspond to the \texttt{actionlib} tasks described just above and another potential operations. The flow can later be defined and re-arranged constantly, creating new behaviors and expanding on what is already done. This \emph{modularity} and \emph{composability} (due to the hierarchical nature of the trees) of BTs with respect to alternatives like FSM is one of the advantages of the formulation \cite{Colledanchise_2018}, and was very relevant for defining the missions in the fleet and adapting to the lessons learned during the rehearsals. Missions could be re-defined very fast, and behaviours could be developed in parallel and easily integrated as sub-trees in more complex missions once well tested and validated.

The behavior of the individual robots of the team in the Challenge are designed using BTs. 
The mission executive that carries out and monitors the mission defined as a BT uses a BT engine. We have employed a publicly available BT implementation\footnote{\url{https://www.behaviortree.dev}}. We have developed a ROS plugin for this engine. It ships with a library containing ROS subscribers, publishers, services and \texttt{actionlib} leaf nodes (mapping adequately the states of these actions to the corresponding states for the nodes BT) to be able to send and receive messages, and call services/actions to/from other nodes running in the network. And it includes a \texttt{BT Mission Execution} ROS node that offers an additional interface (as an \texttt{actionlib} as well) to load, start and stop user-defined behavior trees. This node (see Fig. \ref{fig:actionlib}) runs onboard each robot, providing them with full operational autonomy to carry out complete missions.

\subsection{Cooperation and Coordination}

The approach to Challenge 3 takes advantage of the use of a heterogeneous robot team (mixing UGV and different types of UAVs).
On the one hand, each robot specializes on one of the tasks for the Challenge. This means that we consider a loose cooperation between the robots. In particular, we devote the ground robot for the ground floor indoor fire, and each UAV to a different facade (with the larger M600  for the outside fires and the facade closest to the outside fires). 

The Control Station module (see Fig. \ref{fig:actionlib}) commands the different robots to execute out their missions, and those are carried out autonomously by each robot. Each robot runs all modules onboard, including a different ROS master node. We use the software described in \cite{pound19} to share between robots only vital information through the wireless network, including the position of each robot and the interfaces to control the robots from the station.

Thus, the main aspect to consider is the coordination in time and space to avoid collisions. The task allocation mitigates the risks, as the different vehicles will operate in different facades and, thus, different spaces. In any case, it is important to avoid close encounters between the vehicles. The takeoff zone is very narrow, so the Control Station regulate\revisionDat{s} the order of initiating the missions for each vehicle so that no two UAVs take off at the same time. Furthermore, each vehicle has a landing zone that is separated from the rest of the team.

%% file: 07_ExperimentAndResults.tex
This section presents the results obtained during the 2020 competition the MBZIRC at the National Exhibition Center in Abu Dhabi, (UAE) (see Figure \ref{fig:arena}). \revisionDat{In this section, we focus on the results obtained by the integrated system as a whole, describing a high-level summary of the actions performed by each platform in each event of the competition. For a detail on the performance of the individual systems, please refer to the previous sections.}

The whole integrated multi-robot system was applied 3 times to Challenge 3 during the final event in Abu Dhabi. Twice for the Challenge 3 competition proper (the highest score of both trials \revisionDat{was} kept as the final score), and once more for the Grand Challenge triathlon (please refer to Section \ref{sec:grandchallenge}). The results obtained in the competition are presented in Table \ref{tab:results}. We scored in all trials, always in autonomous mode, and we managed to hit on targets on facade fires and both indoor and outdoor fires. We also managed to autonomously operate a team of 3 robots at the same time. With these results, we achieved the seventh place in Challenge 3 (out of 20 teams in the competition) and the fifth place in the Challenge 3 entry to the Grand Challenge (out of 17) in our first participation in the competition (contributing this way to an overall third place in the Grand Challenge for the team). The results reflect the complexity of the challenge.  
\revisionLuis{A video highlighting results from the MBZIRC competition can be accessed at \url{https://youtu.be/sx9R-6JrfQA}.}

\begin{table}[h]
\centering
\caption{Results obtained during the MBZIRC 2020 competition. All tasks \revisionDat{were} achieved in autonomous mode. Maximum score for a task is by putting 1 litre or covering 100\% of fire.} 
 \begin{tabular}{||p{1.4cm} | p{4cm} | p{4cm} |p{4.2cm}|p{0.7cm}|} 
 \hline 
  & Outdoor Fires with blanket (Weights: 10 UAV, 5 UGV) & Facade Fires with UAVs (Weights: 14 ground floor, 8 first and second floor) & Indoor Fires (Weights: 24 second floor, 16 first floor, 10 ground floor)&Score\\ [0.5ex] 
 \hline\hline
 TRIAL 1 & 1 fire, UAV. 50\% covered & None  &  None  & 5 \\ 
 \hline
TRIAL 2 & None & None & Ground floor, UGV. 350 ml & 3.5 \\
 \hline
 GRAND  & N/A & Ground floor fire, under wind, UAV. 8 ml & Ground floor, UGV. 3 ml & 0.13 \\
 \hline
\end{tabular}\label{tab:results}

\end{table}





\subsection{Trial 1}
\begin{figure}[!t]
	\centering
    \includegraphics[width = 0.9\textwidth]{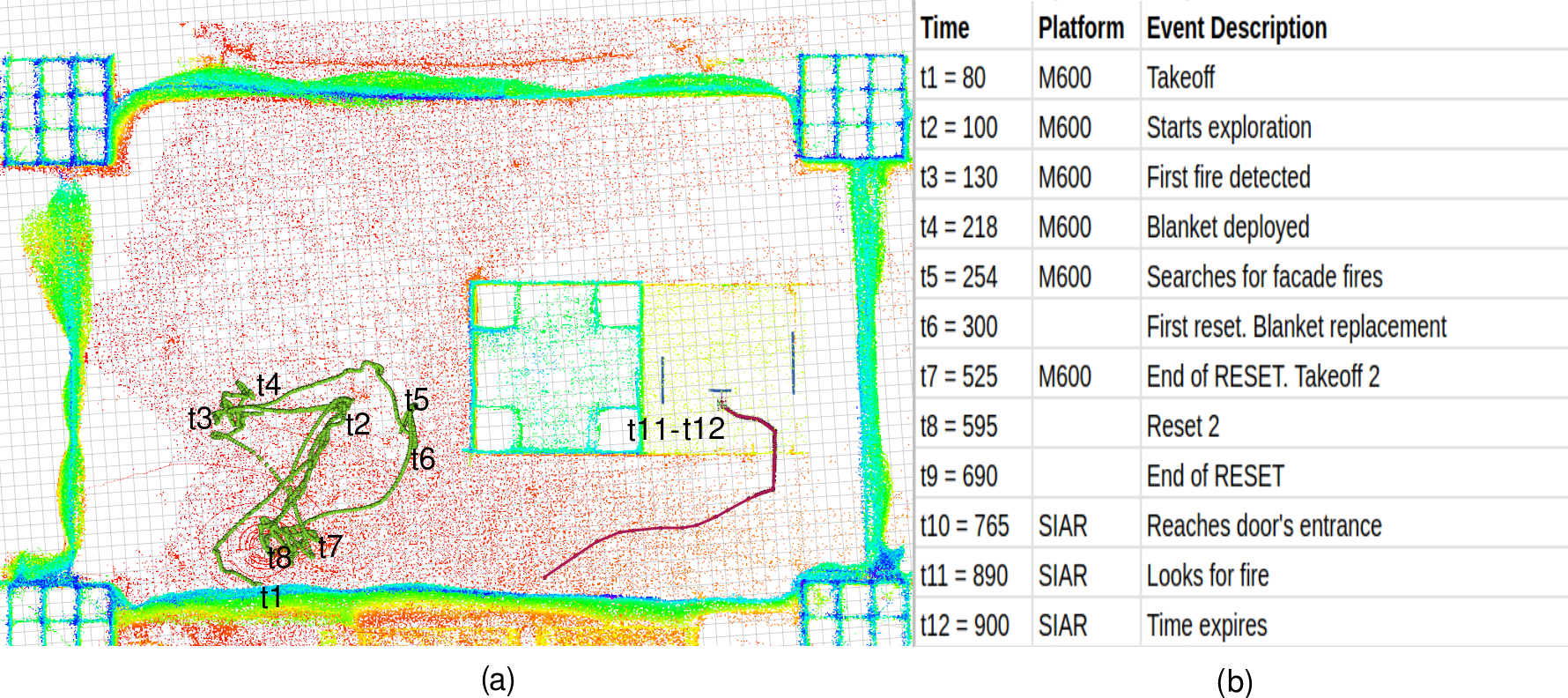}
	\caption{(a) Trajectories of the 2 robots employed in Trial 1. The M600 UAV and the UGV SIAR trajectories are depicted in green and magenta, respectively. The labels indicate the timestamps when the main events occurred. (b) Task performed by each robot with the corresponding timestamp.}
	\label{fig:sequence_trial1}
\end{figure}

\revisionDat{
In this trial the ground robot SIAR was tasked to look for and to extinguish the fire in the ground floor inside the building, while the M600 was tasked to extinguish one outdoor floor with the blanket and then one facade fire with water. We had technical issues that prevented us from flying with the M210 platform. In contrast, our M600 platform did perform very well. It was able to successfully deploy a blanket which partially covered an outdoor fire. 
Figure \ref{fig:sequence_trial1} shows the timeline of the most important events that took place during the first trial.} 

\begin{figure}[!t]
	\centering
    \includegraphics[width = \textwidth]{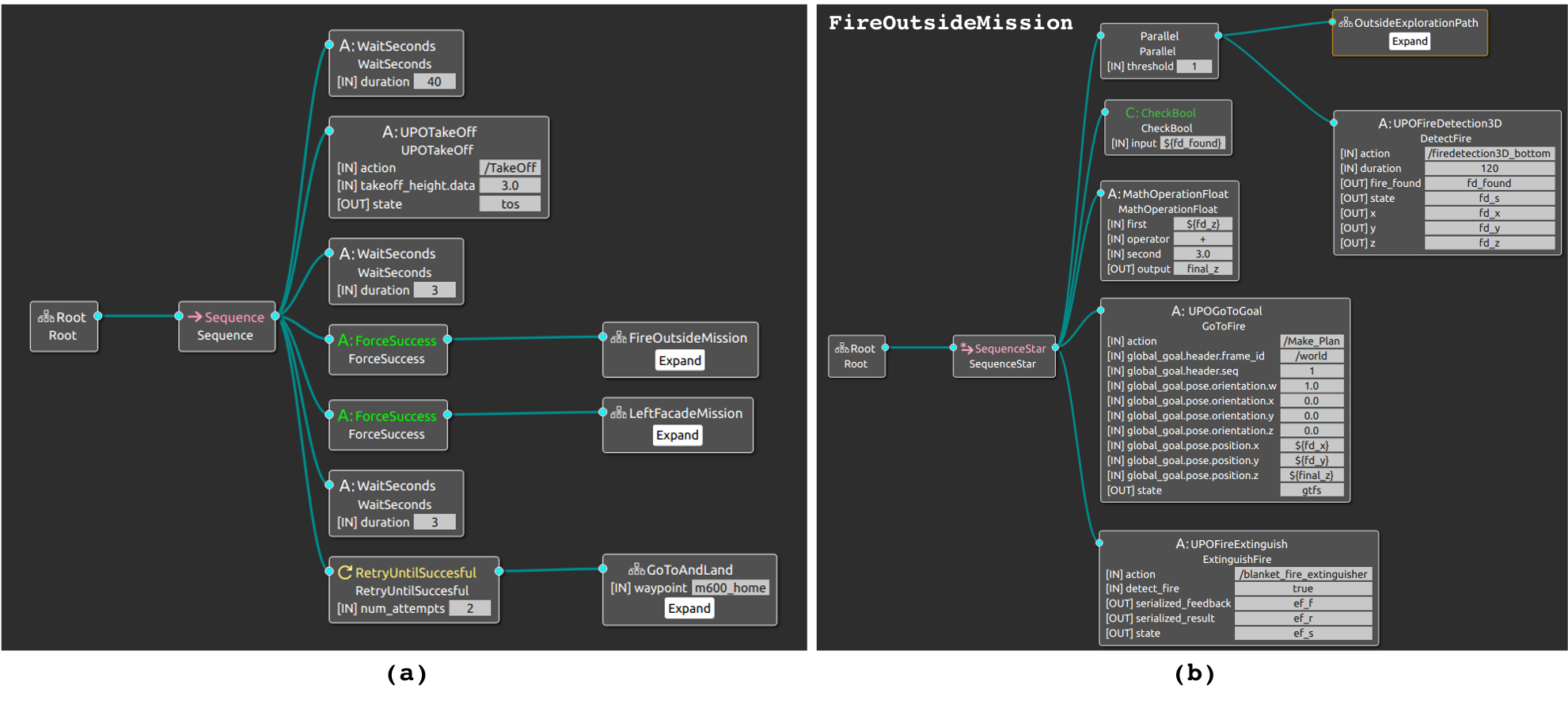}
	\caption{(a) General tree of a (simplified) BT for the M600 mission. The robot is tasked to put out an outside fire and then a fire on the left the facade of the building. The \texttt{Sequence} composite node executes the leafs from top to bottom. If a node returns \texttt{FAILURE}, the whole mission fails. Leaf nodes \texttt{FireOutsideMission}, \texttt{LeftFacadeMission} and \texttt{GotoAndLand} are sub-trees. (b) \texttt{FireOutsideMission} sub-tree.}
	\label{fig:bt-fireexploration}
\end{figure}

\begin{figure}[!t]
	\centering
   \includegraphics[width = 0.9\textwidth]{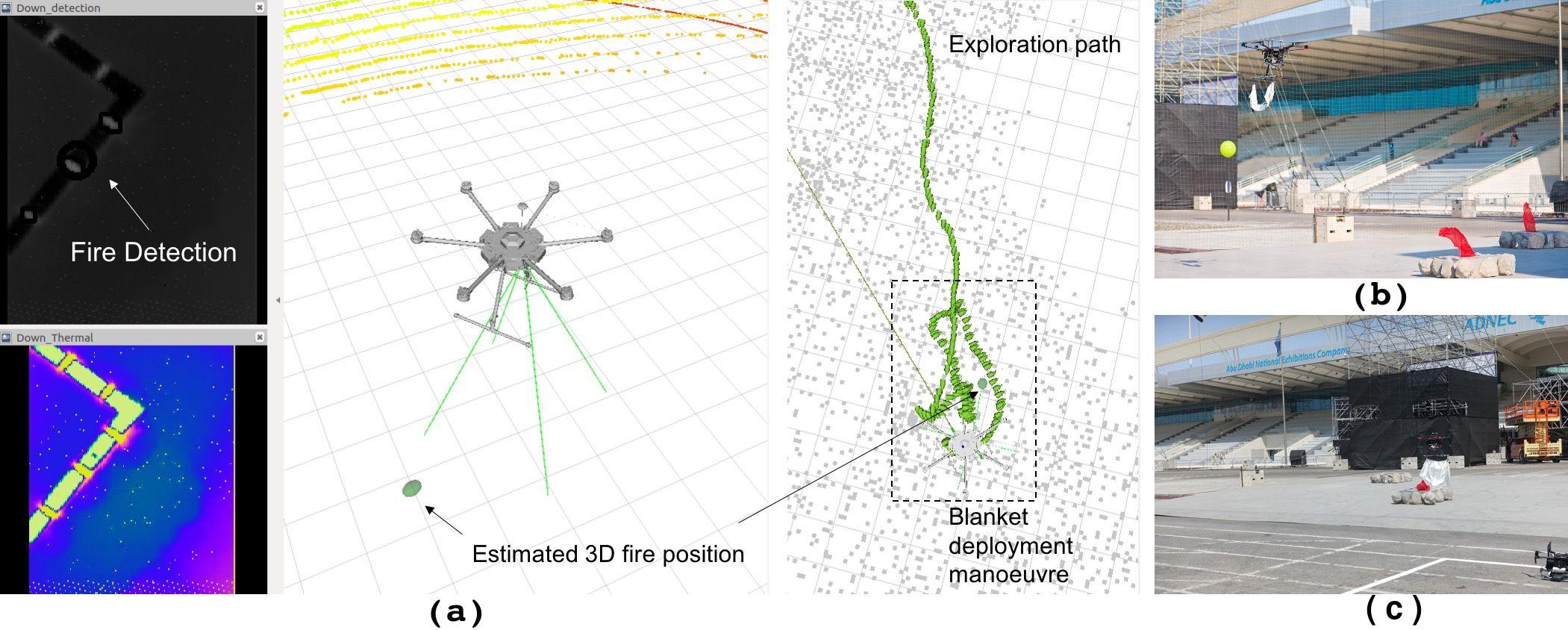}

	\caption{Data from one blanket deployment task of M600 UAV. (a) An exploration path is given and the methods of Section \ref{sec:firedetect} are used to detect and to estimate the 3D position of the fire. Then, the blanket deployment manoeuvre is carried out. (b) Exploration phase. (c) Blanket deployment phase.}

	\label{fig:day2_blanket}
\end{figure}


\revisionLuis{The mission definition for the M600 robot as a BT is presented in Fig. \ref{fig:bt-fireexploration}. The UAV takes off (\texttt{UPOTakeOff} block) after a timeout (to wait for the ground robot to be at a safe distance; \texttt{t1} in Fig. \ref{fig:sequence_trial1}). Then, the fire blanket extinguishing mission proceeds
, represented by the \texttt{FireOutsideMission} sub-tree. Figure \ref{fig:bt-fireexploration} (b), shows the elements of this sub-tree. First, the \texttt{Parallel} composite node executes in parallel a path to explore (\texttt{OutsideExplorationPath}, which uses the navigation actions of Section \ref{sec:navigation}) and activates the \texttt{UPOFireDetection3D} action of Section \ref{sec:firedetect}. Once one of the two nodes returns \texttt{SUCCESS} (either because a fire is detected or the exploration path is finished), the \texttt{Parallel} composite returns \texttt{SUCCESS} (\texttt{FAILURE} otherwise).} 

\revisionLuis{In this case, the exploration path is followed until \texttt{t3} in Fig. \ref{fig:sequence_trial1}. The leaf node \texttt{CheckBool} returns \texttt{SUCCESS} as the variable \texttt{fd\_found} is \texttt{TRUE}. This means that a fire has been detected by the module, and its estimated coordinates are stored at (\texttt{fd\_x}, \texttt{fd\_y}, \texttt{fd\_z}). Then, the UAV is commanded to a fire attack point 3 meters over the estimated position (computed using the \texttt{MathOperationFloat} leaf), by calling to the corresponding \texttt{UPOGoToGoal} (Section \ref{sec:navigation}). Once there, the extinguishing procedure with the blanket is activated (\texttt{UPOFireExtinguish}, Section \ref{sec:fireextblanket}.3). The blanket is successfully deployed at time \texttt{t4}, covering 50\% of the fire.} Figure \ref{fig:day2_blanket} depicts the main steps of the fire blanket mission for the M600 UAV, as described in Fig. \ref{fig:bt-fireexploration}.

\revisionLuis{After the blanket mission, and regardless of its outcome (through the \texttt{ForceSuccess} decorator), the M600 proceeds to the facade fire extinguishing sub-mission. 
This sub-mission (\texttt{LeftFacadeMission}) is very similar to the \texttt{FireOutsideMission} described above (see Fig. \ref{fig:bt-fireexploration}, b)), with different parameters. The main difference is that the exploration path is composed by a set of waypoints that cover the facade, with the UAV oriented towards it. The second difference is that the frontal IR camera is used, and, when a fire is detected, an fire attack waypoint is generated 2.5 meters in front of it. Finally, the \texttt{UPOFireExtinguish} task is called with the specialized version of Section \ref{sec:firefacadewater}.2}.

\revisionLuis{In normal conditions, after the exploration and regardless of the result of the facade mission (through the second \texttt{ForceSuccess}), the robot will go back to the home position. In this case, however, the mission was interrupted at \texttt{t6}, when we asked for a reset to replace the deployed blanket and to start the UGV, instead of waiting for the full exploration of the facade.}

\revisionLuis{After the reset (at \texttt{t7}), the M600 is tasked the same mission with a different exploration path to find the second fire outside, but the blanket was released accidentally just after take off, as the magnetic gripper failed, so a second reset was asked. The UGV SIAR proceeds to its mission (described in detail in Trial 2), but it can only explore one point inside the building (\texttt{t11}) before the trial time runs out.}

\subsection{\revisionDat{Trial 2}}

\begin{figure}[!t]
	\centering
    \includegraphics[scale = 0.27]{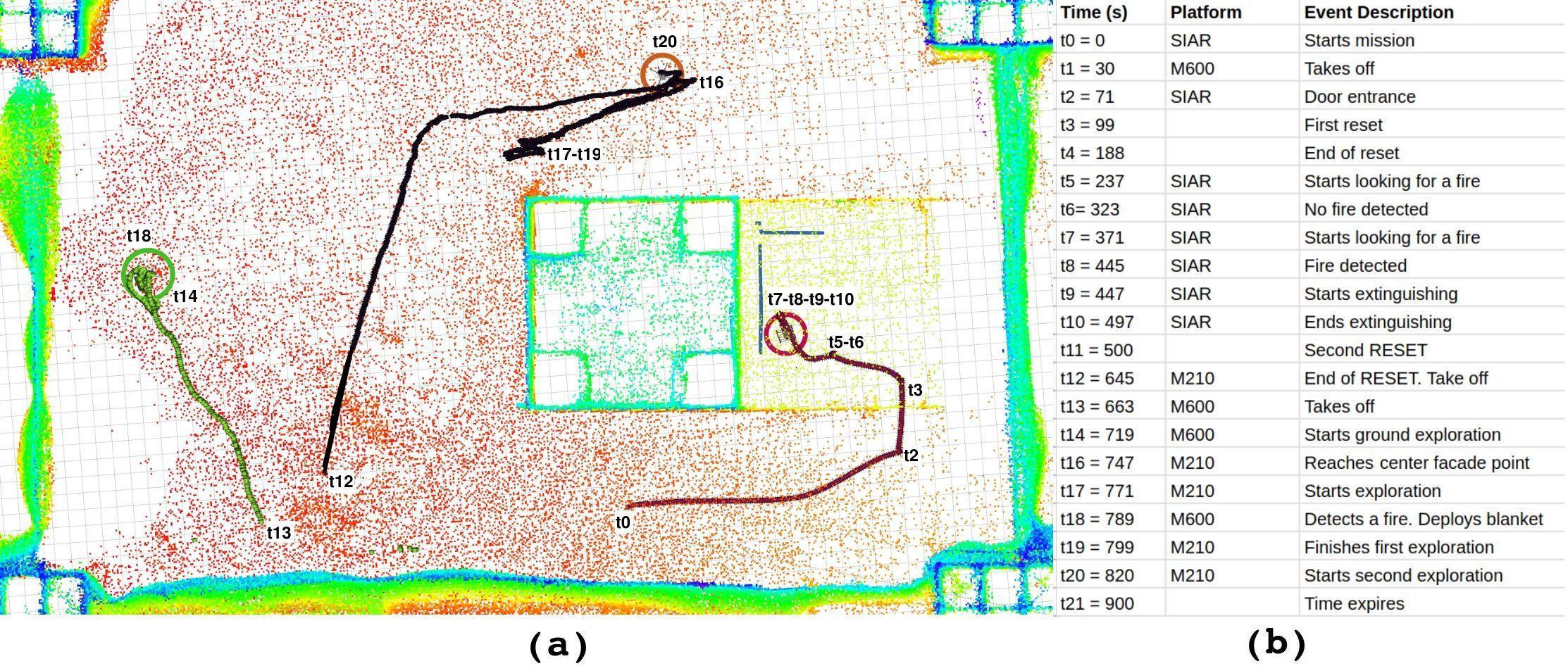}
	\caption{(a) Trajectories of the 3 robots employed in Trial 2. The M600 UAV was tasked with extinguishing an outdoor fire (green trajectory). The M210 UAV was commanded to extinguish a fire in the opposite facade of the building (black trajectory). The UGV SIAR was commanded to look and extinguish the indoor fire in the ground floor (red trajectory). (b) Task performed by each robot with the corresponding timestamp.}
	\label{fig:trial2}
\end{figure}

\begin{figure}[!t]
	\centering
    \includegraphics[width = \textwidth]{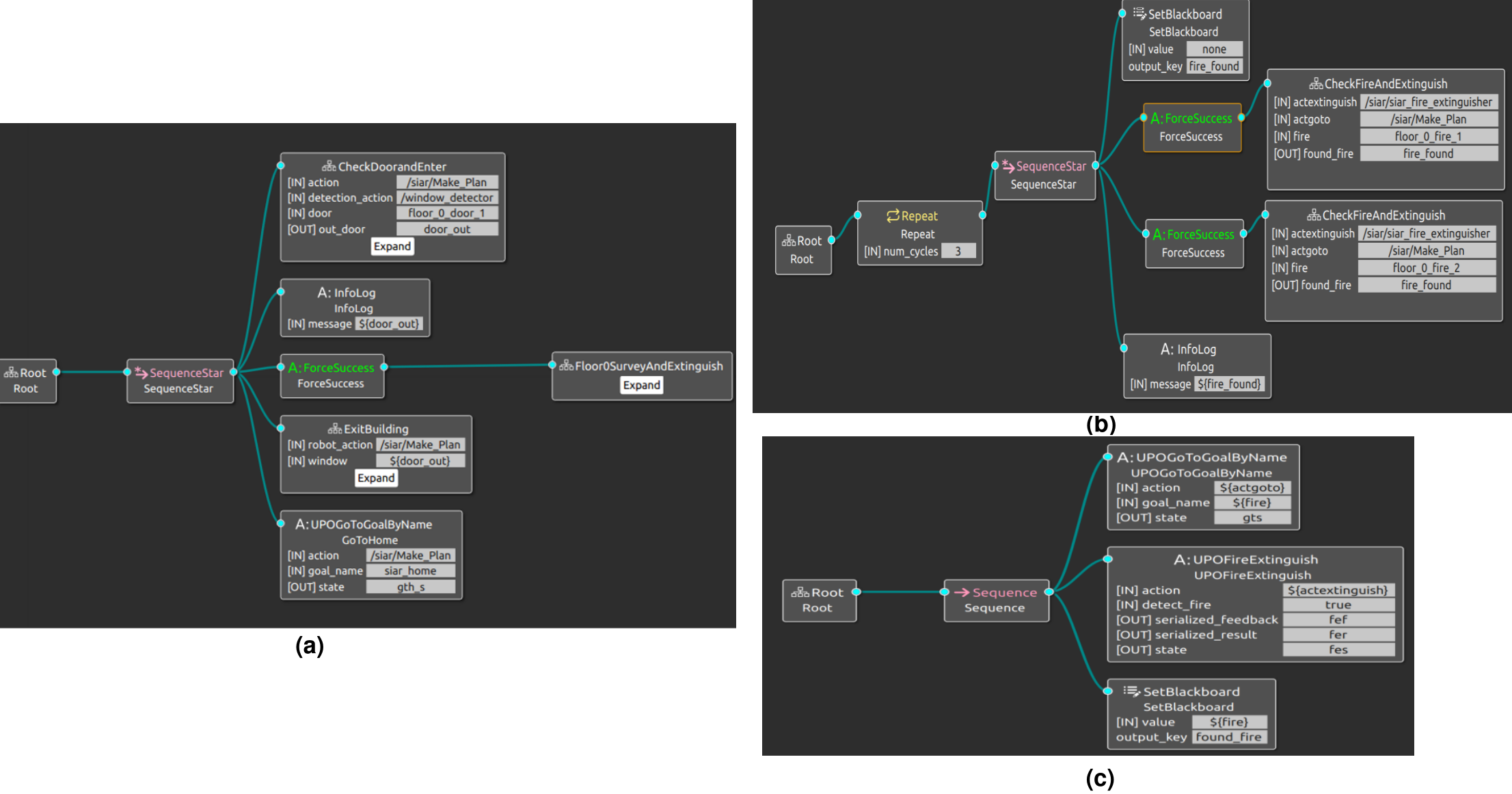}
	\caption{a) General tree of a (simplified) BT for the SIAR mission to put out one fire on the ground floor of the building. b) Subtree \texttt{Floor0SurveyandExtinguish}. c) Subtree \texttt{CheckFireandExtinguish}.}
	\label{fig:bt-siar}
\end{figure}

\begin{figure}[!t]
	\centering
    \includegraphics[width = 0.95\textwidth]{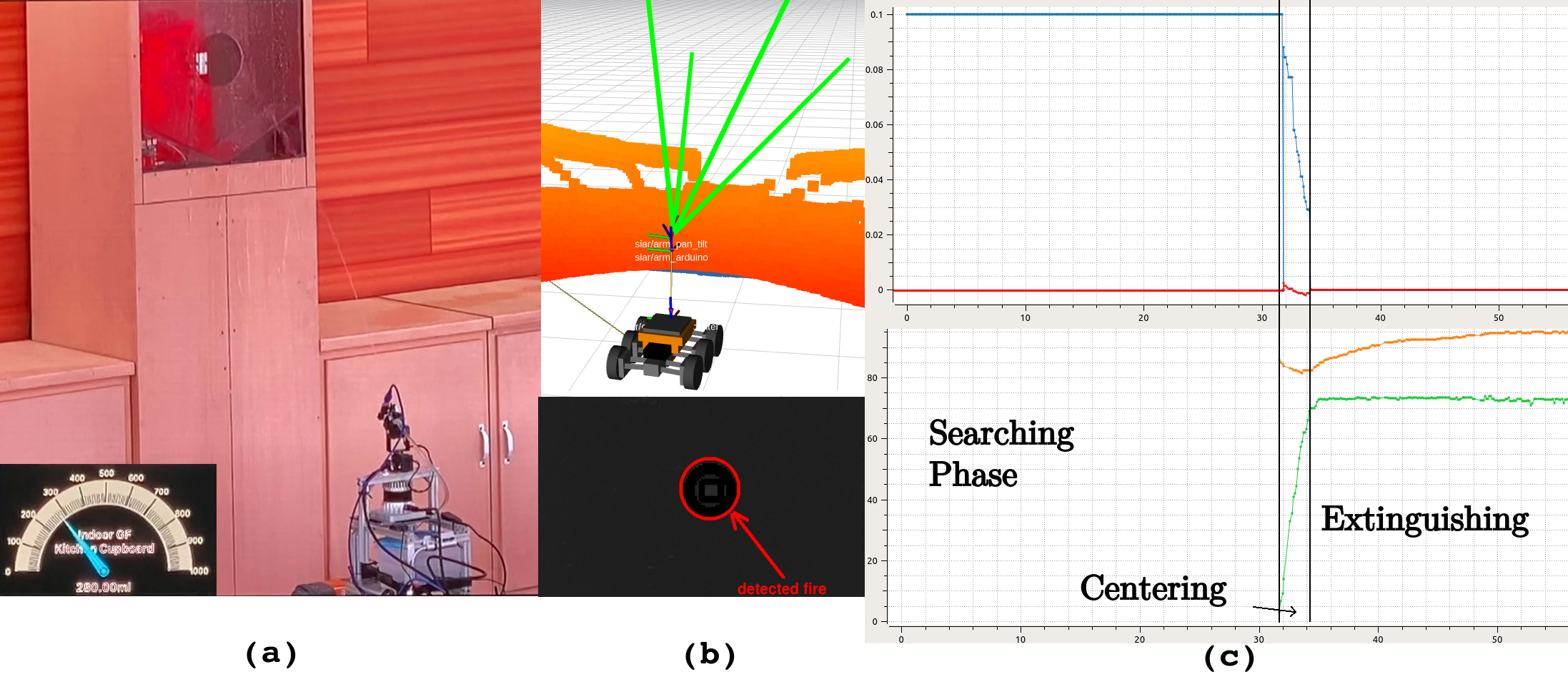}
	\caption{Fire extinguishing action performed by the UGV in Trial 2. (a) Snapshot of the video recorded by the organization in the extinguishing action. The amount of water poured in can be seen below. (b-up) Internal state of the SIAR and FOV of the thermal camera. (b-bottom) Detected fire on the IR image. (c) The upper plot represents the pan and tilt velocity commands on blue and red lines, respectively. In the bottom plot, the image coordinates are plotted in light green and orange, respectively. }
	\label{fig:fire_inside}
\end{figure}

\revisionLuis{In Trial 2, we included the second UAV, M210, that could not participate in Trial 1. As before,} the ground robot was tasked to look for and extinguish the fire in the ground floor, and the M600 was tasked to extinguish one outdoor fire with the blanket and one facade fire with water. \revisionLuis{The M210 was tasked to attack a fire in a different facade of the building}. \revisionDat{Figure \ref{fig:trial2} shows the general evolution of Trial 2.}

\revisionLuis{At the beginning of this trial M600 and SIAR initiated their missions. However, after taking off again the M600 accidentaly dropped its blanket due to a magnet malfunction. A reset was called and then the UGV SIAR was able to carry out its mission successfully. Figure \ref{fig:bt-siar} presents the mission of SIAR. First, the UGV is commanded to enter the building, which occurs at timestamp \texttt{t2} in Fig. \ref{fig:trial2}. Then, the \texttt{Floor0SurveyAndExtinguish} submission is activated. It consists of inspecting two different points using the \texttt{CheckFireAndExtinguish
} sub-tree (see Fig. \ref{fig:bt-siar}, b and c). When a inspection point is reached, the block \texttt{UPOFireExtinguish} is called. The onboard infrared thermal camera checks if a fire exists using the procedure of Section \ref{sec:fireextinguish}.1. If no fire is present, the module returns \texttt{FAILURE} (this occurs, correctly, as there is no fire, in the first inspection point between times \texttt{t5} and \texttt{t6})}. \revisionDat{Otherwise, first a centering action is performed and finally the fire extinguishing system is activated to deploy 1 litre (see Fig. \ref{fig:fire_inside}, a). This actually occurs in the second inspection point (between times \texttt{t9} and \texttt{t10}). Figure \ref{fig:fire_inside} shows the UGV extinguishing this indoor fire during Trial 2. We are able to apply 350 ml on target from the whole liter, because of a small misalignment of the water ejector and the thermal camera}. The mission is not designed \revisionDat{to deploy all the water on the UGV, allowing it} to repeat exploring the environment for more fires, to compensate for potential false alarms. This turned out to be too conservative, as the system was actually capable of detecting the real fire present and no false alarms were created. 






\revisionLuis{After the UGV put out the inside fire, a new reset is called (in time \texttt{t11}) to initiate the UAVs. The M210 takes off first at \texttt{t12} and is sent to explore the farthest facade. 20 seconds later, at \texttt{t13}, the M600 takes off.} \revisionLuis{The M600 is tasked the same mission as in Trial 1, with a change in the order of the exploration path waypoints to add a safety buffer with the M210 UAV. In this case, the outdoor fire is again correctly detected and located, and the blanket is deployed (at \texttt{t18}), but misses the target by a few centimeters. We actually increased the height at which the deployment manoeuvre was performed from Trial 1 to 2 to add some safety buffer, and this ends up being detrimental. There is not further time to carry the facade part of the M600 mission.}

\revisionLuis{The M210 is able to explore part of the facade looking for the fires (\texttt{t16} until \texttt{t21}). However, we run out of time during Trial 2 
before it can detect any fire. Please notice that this mission is the same as the M600 facade sub-mission described above, changing the exploration path waypoints. This shows one of the main advantages of using BTs: its modularity allows us to reuse BTs from different platforms with minimal changes.}

\subsection{\revisionDat{Grand Challenge}}\label{sec:grandchallenge}

\begin{figure}[!t]
	\centering
    \includegraphics[width = 0.9\textwidth]{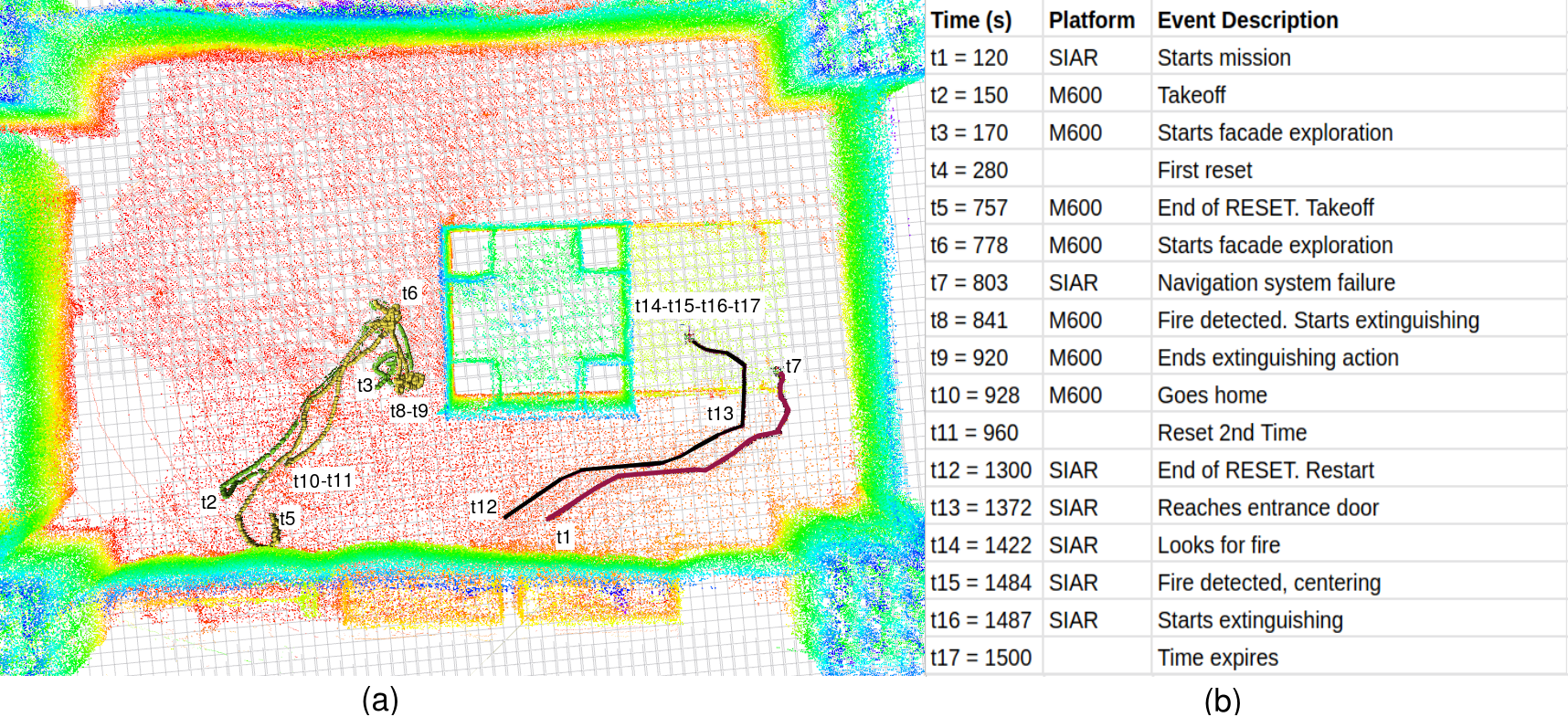}
	\caption{(a) Trajectories of the M600 (green and yellow) and SIAR (purple and black) in their first and second missions, respectively. (b) Task performed for each robot with the corresponding timestamp.}
	\label{fig:timelapse_grandchallenge}
\end{figure}

\revisionLuis{In the Grand Challenge, a reduced version of the three MBZIRC Challenges are carried out at the same time. Regarding Challenge 3, the fires to put out with a blanket are eliminated. There is also a limit in the number of robots for tackling all challenges, and thus we could only use our M600 platform and the UGV. 
The total amount of time was raised from 15 to 25 minutes, but resets are global to all challenges.}


\revisionDat{Figure \ref{fig:timelapse_grandchallenge} shows the main events that occurred during the Challenge 3 entry to the Grand Challenge. 
The mission for the M600 drone is a stripped down version of the mission in Trials 1 and 2, eliminating the blanket deployment (again, this was very easily carried out given the composition of the associated BT). In contrast to the experiments in Trial 2, in this case our M600 drone manages to partially extinguish a facade fire. After a first reset at \texttt{t4} (due to other Challenge), it follows an exploratory path until it detects an active fire with its thermal camera on front at \texttt{t8}. As soon as it detects the fire, it switches into fire centering mode and starts to eject water to extinguish it (see Fig. \ref{fig:control_facade_m600}).} \revisionLuis{This particular fire is given larger weights for the scoring system as it was subjected to lateral wind gusts of up to 8 m/s.}

\revisionDat{Regarding the UGV platform, the mission is the same as before, with the main change of deploying the whole tank of water and returning to home position instead of searching for the remaining fires. 
It does not perform as well as in the previous day as the navigation system gets stuck when trying to enter in the building at timestamp \texttt{t7}. Since requesting a reset affects and stops every challenge, it is necessary to wait for the opportune moment and coordinate the whole team to intervene the platform and restart its system.} 

\revisionDat{Later, at time \texttt{t12} we reinitialize the UGV 
and thus finally the second mission is successfully carried out. The UGV heads to the first inspection point, and then it executes the searching phase (Section 5.1) looking for fire at time \texttt{t14}. 
As soon as it detects the fire (located at a different point than the previous day), the centering action is executed (\texttt{t15}) and the UGV is able to partially extinguish the fire on the ground floor. Unfortunately, the time expires only a few seconds after the start of the extinguishing procedure (\texttt{t16}), and thus only a small quantity of water is deployed on target.
}

%% file: 10_ConclusionsNew2.tex
We present in this article our entry to tackle MBZIRC Challenge 3: ``Team of Robots to FightFire in High Rise Building",  using cooperative multi-robot team framework approach. The system presented achieved the 7th place (same score as 6th, longer mission time) in Challenge 3 and contributed to the 3rd place in the Grand Challenge \revisionLuis{of the team}. The difficulty of this Challenge can be seen in that only 9 teams were able to score in autonomous mode between the two trials of Challenge 3, and only 5 teams during the Grand Finale trial of the same Challenge 3. We managed to score in autonomous mode in all three occasions. 

\revisionLuis{And while the system and techniques presented in the paper were developed with the MBZIRC challenge in mind, we believe the localization, navigation and mission executive can be adapted and applied to other urban scenarios.}

At the same time, the system was not able to perform as robustly as expected and evidently some other teams did better. Some of the lessons learned are presented here:

\revisionLuis{\textbf{Hardware matters}}: Being a robotics competition means that not only software matters, but hardware selection and design decisions are key. The use of off-the-shelf platforms was a good decision, in particular for the drones. The DJI platforms offered a ready to fly system with direct ROS integration. This opened the door to focus on the software development tools required for the robot autonomy, significantly reducing the issues and problems derived from hardware integration in drones. It is also worth to mention that we experimented DJI software issues before and during the competition: onboard firmware mismatch problems, random GPS no-fly area restrictions and motor overheating. Nevertheless, all in all, we consider that the advantages of using a commercial off-the-shelf drone platform was a good decision and that, in general, is preferable to avoid the design/implementation of drones in order to focus in the autonomy problem. The platforms were adapted for the water ejection and blanket deployment systems. \revisionLuis{The results could have been improved by more robust hardwar adaptation. While all in all the water ejection mechanisms were adequate, achieving a higher pressure would have helped in being more accurate}. Regarding the blanket deployment mechanism, while we were able to deploy the blanket on the fire once, \revisionLuis{it was affected by the air turbulence generated by the own UAV which led to some blanket drops, as described above.} 
Finally, our ground robot used a pan and tilt unit instead of an arm like most teams for budgetary and weight restrictions. The use of an arm clearly helps in the final extinguishing manoeuvre.

\revisionLuis{\textbf{Versatile localization}}: Using a map-based localization system allowed us a lot of versatility, even permitting us finally discarding the use of GPS for localization. This approach allowed us for fast system deployment, so that we do not need a precise GPS localization to localize the robot into the map, just the minimum required by the DJI drones to perform the take-off maneuver safely. The localization had a stable performance and a good precision during the competition, relying only on the use of LIDAR measures that were matched against a 3D map obtained a priori. It also was very important in order to be able to safely navigate in the scenario, as the localization of our platforms was performed taking into account the different elements on the stage, e.g. columns or safety nets, allowing us to locate them with good precision against the obstacles in the environment. The localization supported the robot autonomous navigation in the scenario, and the map-based system made the indoor-outdoor transition on localization seamless. However, only the ground robot entered the building. Even though we performed missions with the drones inside the building in simulation, finally we did not attempt to enter the building during the competition. \revisionLuis{This was one of the most challenging tasks and, }actually, no team was able to score inside the building during the competition, and only some manual trials were performed. This clearly requires techniques with a higher level of maturity.

\revisionLuis{\textbf{Local operational autonomy}}: Robot local autonomy was also key. Each robot was able to carry out its mission with all processing carried out on board, with the central station only in charge of launching the local missions of the robots in adequate order and relaying some global information. Even under communication dropouts, the robots were able to carry out their missions. The use of BTs as a mission executive on board the robots offered us also a great deal of flexibility to design and compose the missions on the base of the atomic tasks and sub-behaviours already defined. The BT engine also allowed the robots to react to the events detected during the runs. 


\revisionLuis{\textbf{Rapid and robust deployment}}: At the same time, robustness is fundamental to operate several robots at the same time. During trials, preparation time was very short. While we were able to perform missions with 3 robots at the same time, \revisionLuis{the system start-up was not robust enough} to fully leverage the 15 minutes of the missions. \revisionLuis{While the system can operate under communication dropouts, communication with the robots was needed to start them} and, some times, delays \revisionLuis{in the starting sequence caused by communication problems} prevented to achieve more tasks in the mission. However, we feel that we are in the right direction and we look forward to subsequent editions to improve the robustness and accuracy of our systems and to, hopefully, win the challenge!
